\documentclass[10pt,twocolumn,letterpaper]{article}

\usepackage[pagenumbers]{cvpr}

\usepackage{cuted}
\usepackage{adjustbox}
\usepackage{array}
\usepackage{colortbl}
\usepackage{makecell}
\usepackage{multirow}
\usepackage{pgfplots}
\usepgfplotslibrary{groupplots}
\usetikzlibrary{calc}
\pgfplotsset{compat=1.18}

\graphicspath{{figures/teaser/}}
\definecolor{teaserEgo}{HTML}{58B7FF}
\definecolor{teaserRemote}{HTML}{FFB74D}
\definecolor{teaserCandidate}{HTML}{55D6BE}
\definecolor{teaserOutput}{HTML}{FF5D73}
\definecolor{teaserGT}{HTML}{687386}
\definecolor{tableGain}{HTML}{E8F5E9}
\definecolor{tableDrop}{HTML}{FDECEC}
\definecolor{tableAccent}{HTML}{D6E9FF}
\definecolor{tableMuted}{HTML}{F5F6F7}
\newcommand{\tabgain}[1]{%
    \cellcolor{tableGain}\textcolor{ForestGreen}{\bfseries #1}%
}
\newcommand{\tabdrop}[1]{%
    \cellcolor{tableDrop}\textcolor{BrickRed}{#1}%
}
\newcommand{\tabna}{\textcolor{gray}{\textit{N/A}}}

\makeatletter
\renewcommand\paragraph{\@startsection{paragraph}{4}{\z@}%
    {1.5ex \@plus .5ex \@minus .2ex}{-1em}{\normalfont\normalsize\bfseries}}
\makeatother

\definecolor{cvprblue}{rgb}{0.21,0.49,0.74}
\usepackage[pagebackref,breaklinks,colorlinks,allcolors=cvprblue]{hyperref}

\title{GeoScaffold: From Latent to Realized Geometry in Collaborative Detection}
\author{%
Kang Yang\textsuperscript{1}\quad
Yance Fang\textsuperscript{1}\quad
Gaochen Cui\textsuperscript{2}\quad
Xiaoqiang Jin\textsuperscript{3}\quad
Deying Li\textsuperscript{1}\quad
Yongcai Wang\textsuperscript{1}\\[3pt]
\textsuperscript{1}Renmin University of China\qquad
\textsuperscript{2}CFINS, Tsinghua University\qquad
\textsuperscript{3}QiYuanLab}

\begin{document}
\maketitle
\begin{strip}
    \vspace{-1.5em}
    \centering
    \captionsetup{type=figure,font=small}
    \includegraphics[width=\textwidth]{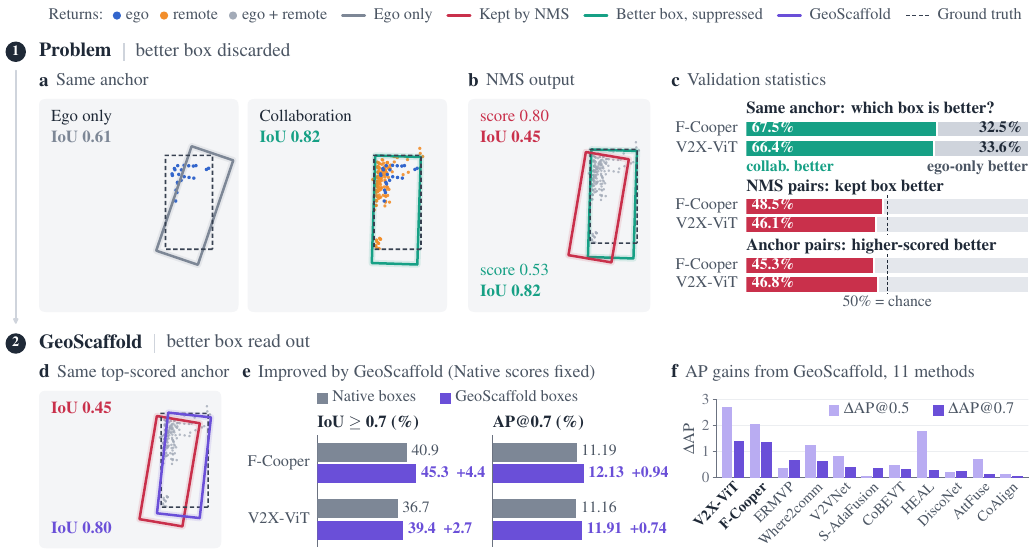}
    \begin{subfigure}{0pt}\phantomsubcaption\label{fig:teaser-observations}\label{fig:teaser-same-anchor}\end{subfigure}%
    \begin{subfigure}{0pt}\phantomsubcaption\label{fig:teaser-readout}\end{subfigure}%
    \begin{subfigure}{0pt}\phantomsubcaption\label{fig:teaser-nms-population}\end{subfigure}%
    \begin{subfigure}{0pt}\phantomsubcaption\label{fig:teaser-geo-case}\end{subfigure}%
    \begin{subfigure}{0pt}\phantomsubcaption\label{fig:teaser-fixed-winner}\end{subfigure}%
    \begin{subfigure}{0pt}\phantomsubcaption\label{fig:teaser-cross-model}\end{subfigure}%
    \vspace{-0.4em}
    \caption{\textbf{Score-based selection often discards better-localized boxes; GeoScaffold
    improves the box at the top-scored anchor.}
    \subref{fig:teaser-observations}, \subref{fig:teaser-readout}, \subref{fig:teaser-geo-case}:
    An illustrative F-Cooper example drawn to the same scale.
    \subref{fig:teaser-nms-population}, \subref{fig:teaser-fixed-winner},
    \subref{fig:teaser-cross-model}: Statistics on the UrbanIng-V2X validation split
    (App.~\ref{sec:collaboration-selection}).
    \subref{fig:teaser-observations}: The same positive anchor under ego-only and collaborative
    inputs.
    \subref{fig:teaser-readout}: NMS keeps the top-scored box (red) and suppresses the
    better-localized collaborative box (teal).
    \subref{fig:teaser-nms-population}: Collaboration improves localization at about two-thirds of
    positive anchors, yet for two boxes of one object, NMS and classification scores pick the one
    with the higher IoU less than half of the time.
    \subref{fig:teaser-geo-case}: Boxes of the original detector (Native, red) and GeoScaffold
    (violet) at the top-scored anchor.
    \subref{fig:teaser-fixed-winner}: With Native scores fixed, GeoScaffold's boxes raise the share
    of objects whose top-scored anchor yields a box with IoU $\geq$ 0.7 and improve AP@0.7
    (App.~\ref{sec:readout-analysis}).
    \subref{fig:teaser-cross-model}: AP gains for all 11 fusion methods when GeoScaffold's scores
    and boxes are both used (Table~\ref{tab:main-results}).}
    \label{fig:teaser}
\end{strip}
\begin{abstract}
Most collaborative 3D detectors fuse bird's-eye-view (BEV) features from connected vehicles and
roadside infrastructure to aggregate complementary observations. Although modern communication,
alignment, and fusion pipelines encode rich latent geometry in the fused BEV features, the
standard readout pipeline inherited from single-agent detectors, i.e., anchor scoring, box
decoding, and non-maximum suppression (NMS), does not fully exploit these geometric cues. We
empirically identify and quantify this gap on UrbanIng-V2X. Collaboration improves localization at
about two-thirds of positive anchors, yet when NMS resolves two overlapping collaborative boxes
of one object, it keeps the better-localized box less than half of the time. A key cause is a
supervision mismatch: classification scores are trained only for foreground-background
discrimination, so nothing in training requires them to rank boxes by localization quality. To
address this limitation, we introduce GeoScaffold, a lightweight training-only auxiliary
objective. It predicts a foreground mask at each positive anchor, i.e., the cells of a grid over
the anchor that lie inside the ground-truth box, to supply the dense spatial supervision that the
seven-dimensional box code leaves implicit. Its loss, added to the detection loss, updates the
encoder and fusion modules through the fused BEV and trains the detector toward a
better-localized box at every positive anchor, including the top-scored one. The auxiliary head
is removed at inference, preserving the original detector architecture, communication logic, and
inference cost. GeoScaffold raises AP@0.5 and AP@0.7 by 0.98
and 0.55 points on average across all 11 fusion methods on the UrbanIng-V2X validation split,
with gains extending to most methods on an unseen intersection and to all ten methods that
produce detections on V2V4Real.
\end{abstract}
\section{Introduction}
\label{sec:introduction}

Most collaborative 3D detectors fuse the bird's-eye-view (BEV) features of connected vehicles and
roadside infrastructure, so that one agent can recover objects that are occluded or truncated in
its own view~\cite{collaborative-survey1,collaborative-survey2,collaborative-survey3}.
Research on collaborative perception has mainly concentrated on what to communicate, how to align
the received features, and how to fuse
them~\cite{Who2comm,when2comm,where2comm,CodeFilling,localizationproblem,Coalign,CoBEVFlow,Fcooper,v2vnet,v2x-vit,CoBEVT,disconet,HEAL},
which together govern how much complementary evidence about an object's position, extent, and
orientation reaches the fused BEV. We call this evidence \emph{latent geometry}, which resides in
the fused BEV until the detector's readout turns it into object boxes. This readout is typically
inherited from single-agent detectors, which score every anchor, regress a compact box code,
decode boxes, and remove duplicates with non-maximum suppression
(NMS)~\cite{fasterrcnn,voxelnet,SECOND,PointPillars,neubeck2006efficient}. This inherited pipeline
implicitly assumes that the latent geometry is realized in the output boxes. Yet this assumption
has rarely been tested: how much of the latent geometry does the readout realize?

Simply comparing the average precision (AP) of collaborative and ego-only
detectors~\cite{OPV2V,dair-v2x,v2v4real} cannot answer this question: a small gain may mean either
that the fused BEV holds little latent geometry or that the readout fails to realize it. To
separate the sources of the collaborative gain, we fix the set of positive anchors, which are the
anchors that the detector's assignment rule matches to a ground-truth object, and vary only the
input, i.e., the full collaborative input or an ego-only variant with all collaborative data
removed. At each positive anchor, the same detector decodes one box from each input.

On the UrbanIng-V2X validation split~\cite{urbaningv2x2025}, the collaborative box has the higher
intersection-over-union (IoU) at 67.5\% of positive anchors under F-Cooper~\cite{Fcooper} and
66.4\% under V2X-ViT~\cite{v2x-vit} (Figs.~\ref{fig:teaser-observations}
and~\ref{fig:teaser-nms-population}). These rates show that the latent geometry added by
collaboration does reach most positive anchors. When NMS resolves two overlapping collaborative
boxes of one object, however, it keeps the better-localized one in only 48.5\% (F-Cooper) and
46.1\% (V2X-ViT) of cases (Figs.~\ref{fig:teaser-readout} and~\ref{fig:teaser-nms-population}).
This box selection has a measurable cost: the best-localized box across an object's positive
anchors meets the IoU $\geq$ 0.7 criterion for 48.3\% of objects under F-Cooper and 44.1\% under
V2X-ViT. By contrast, the box at the top-scored positive anchor, which NMS favors, satisfies this
threshold for only 40.9\% and 36.7\% of objects, respectively
(App.~\ref{sec:collaboration-selection}; Fig.~\ref{fig:teaser-fixed-winner}).

This issue arises from the ranking rule: candidate boxes are ordered by classification scores,
yet all positive anchors assigned to one object share a single classification label.
Consequently, training imposes no constraint that forces classification scores to correlate with
localization quality. This supervision mismatch is well documented in single-agent detection,
where IoU-aware scoring and IoU-guided NMS align box selection with localization
quality~\cite{IoUNet,GFL,VFL,CIASSD}. We demonstrate that this mismatch still exists in
collaborative fusion. For pairs of positive anchors belonging to the same object, the
higher-scored anchor yields the box with the higher IoU in only 45.3\% (F-Cooper) and 46.8\%
(V2X-ViT) of cases, a rate no better than random ordering (Fig.~\ref{fig:teaser-nms-population}).
The practical impact of this mismatch depends on the localization quality of the
boxes being ranked. If every positive anchor predicts a well-localized box, any box retained by the selection
procedure is well localized. Motivated by this observation, we aim to train the detector to
produce accurately localized boxes at all positive anchors.

Each positive anchor's box is regressed from the fused BEV around the anchor, but the
localization loss of the original detector (Native) supervises it only through a
seven-dimensional box code~\cite{SECOND,PointPillars}, which constrains seven numbers per anchor
and leaves implicit which parts of the anchor the object covers. We propose GeoScaffold to
explicitly supply this dense spatial coverage supervision. Like scaffolding on a building site,
it shapes the fused representation during training and is removed once training ends.
Specifically, at each positive anchor of the detector's own assignment, GeoScaffold uses a
lightweight head to sample the fused BEV in an anchor-aligned frame and to predict the anchor's
foreground mask: which cells of a grid over the anchor lie inside its assigned ground-truth box.
Its loss is added to the detection loss and updates the fusion module and the encoder through the
fused BEV. Because the detection head regresses each positive anchor's box from this fused BEV,
GeoScaffold trains the detector toward a better-localized box at every positive anchor.

GeoScaffold adopts the native anchor assignment strategy, so its supervision applies to exactly
the same positive anchors used by the original detection losses, including each object's
top-scored positive anchor, which NMS favors. Since it only takes fused BEV features and anchor
assignments as input, GeoScaffold can be plugged into any anchor-based BEV fusion detector,
regardless of the specific fusion pipeline. During inference, the whole detection pipeline
remains unchanged: communication, feature fusion, scoring, box decoding, and NMS are unmodified,
with no extra parameters or computational overhead.

If GeoScaffold realizes latent geometry by improving box predictions, it should improve
localization at the Native detector's top-scored positive anchors even when the scores remain
unchanged. To isolate the effect of box quality from score changes, we fix the Native
classification scores of F-Cooper and V2X-ViT, which keeps each object's top-scored positive
anchor identical, and replace only the decoded boxes with GeoScaffold's. Under this controlled
setup, the class-balanced fraction of objects whose top-scored positive anchor yields a box with
IoU $\geq$ 0.7 increases from 40.9\% to 45.3\% for F-Cooper and from 36.7\% to 39.4\% for
V2X-ViT (Figs.~\ref{fig:teaser-geo-case} and~\ref{fig:teaser-fixed-winner}). Beyond this test,
GeoScaffold consistently improves AP@0.5 and AP@0.7 across all 11 fusion methods on the
UrbanIng-V2X validation split, with average gains of 0.98 and 0.55 points, respectively
(Fig.~\ref{fig:teaser-cross-model}). These improvements extend
to most fusion methods on the held-out crossing3 intersection and to all ten methods that produce
detections on V2V4Real~\cite{v2v4real}.

Our contributions are as follows:
\begin{itemize}
    \item We identify and quantify a gap between latent geometry and realized boxes in
    collaborative 3D detection, and trace it to score-based box selection.
    \item We propose GeoScaffold, a training-only objective that predicts a foreground mask at
    each positive anchor, supplying the dense spatial supervision that the seven-dimensional box
    code leaves implicit, at no additional inference cost.
    \item GeoScaffold raises AP@0.5 and AP@0.7 for all 11 fusion methods on the UrbanIng-V2X
    validation split, and the gains extend to most methods on a held-out intersection and to all
    ten methods that produce detections on V2V4Real, suggesting that unrealized latent geometry
    is common across scenes, sensor platforms, and fusion designs.
\end{itemize}
\section{Related Work}
\label{sec:related-work}

\subsection{Collaborative Perception}

Collaborative perception has progressed from sharing point clouds or detection
outputs~\cite{early1,early2,late1} to fusing learned BEV features, as in the 11 fusion methods
we evaluate (Table~\ref{tab:main-results}). Learnable handshakes and communication graphs decide
which agents interact~\cite{Who2comm,when2comm}, and instance- and query-level methods make the
exchanged content selective~\cite{TransIFF,quest}. Sparsification, quantization, codebooks, and
reconstruction of omitted content reduce
bandwidth~\cite{ACCO,CodeFilling,how2comm,diffcp,EIMC,sparsealign}. Robustness-oriented methods
correct pose error~\cite{localizationproblem,ROCO}, compensate for latency and asynchronous
sensing~\cite{latency1,CoBEVFlow,Traf-align}, tolerate lossy communication~\cite{latency2},
denoise shared features~\cite{MRCNet,codiff}, defend against malicious
agents~\cite{ROBOSAC,MADE}, estimate uncertainty~\cite{conformal_uncertainty}, or exploit
spatiotemporal context~\cite{RobustMulti-AgentCollaborativePerceptionviaSpatio-TemporalAwareness}.
Heterogeneous methods align agents with different sensors, models, or task
interfaces~\cite{HM-VIT,bm2cp,v2vformer++,cocmt,HEAL,stamp,PHCP,PolyInter,congzhangnegocollab,GenComm,yang2026bolt}.
Collaboration also improves camera-based 3D detection~\cite{CoCa3D}, semantic
segmentation~\cite{GSCOOP}, scene completion~\cite{STAR}, and semantic occupancy
prediction~\cite{CoHFF}. Simulated~\cite{V2XSim,DOLPHINS} and real-world
benchmarks~\cite{dair-v2x,v2v4real,v2x-seq,RCooper,TUMTrafV2X,v2xreal,urbaningv2x2025} cover
vehicle--vehicle, vehicle--infrastructure, infrastructure--infrastructure, and mixed-agent
collaboration, and some
also cover sequential perception and forecasting. U2UData~\cite{U2udata} extends collaborative
perception to aerial platforms. For detectors that fuse BEV features, communication,
alignment, and fusion determine how much geometric evidence about each object reaches the fused
BEV. GeoScaffold targets the step that follows: it trains the detector so that its readout turns
more of this latent geometry into well-localized boxes.

\subsection{Box Readout and Geometry Supervision}

3D detectors decode objects from voxels, pillars, points, or BEV maps through anchors, centers,
votes, or geometric
primitives~\cite{voxelnet,SECOND,PointPillars,Pointrcnn,Pvrcnn,Center-based,VoteNet,ThreeDSSD,H3DNet,HotSpotNet,FCAF3D}.
Most rank the decoded boxes by classification scores before NMS; IoU-aware
scores~\cite{GFL,VFL,CIASSD} and IoU-guided suppression~\cite{IoUNet} align this ranking with
localization quality.

Dense targets supplement box supervision. Mask heads trained with the box head on each region of
interest (ROI)~\cite{He_2017_ICCV,PointRend} raise box AP in Mask R-CNN, and box-derived masks improve
2D detection~\cite{Brazil_2017_ICCV,Zhang_2018_DES} and train instance
segmentation~\cite{BoxInst}. In 3D, box-derived point labels supervise
detectors~\cite{Pointrcnn,Shi_2021_PartA2,He_2020_CVPR}, as do shape
targets~\cite{DOPS,RfDNet,PGRCNN}, and further methods add geometric cues or dense occupancy
supervision~\cite{PARTNER,BEVFormerV2,GeoMIM,BEVAtDC,Zhang_2024_CVPR,TPVFormer,OccFormer,SurroundOcc,OccNet,COTR,PanoOcc,OccupancyM3D}.
Auxiliary branches can be discarded after
training~\cite{He_2020_CVPR,Mostajabi_2018_CVPR,AuxiliaryTraining}.
In collaborative perception, DiGS-CP~\cite{DiGS-CP} adds generative supervision that
reconstructs object-level point-cloud representations. GeoScaffold brings box-derived dense
supervision to collaborative detection and attaches it to the detector's own positive anchors,
whose boxes the readout ranks.

\begin{figure*}[t]
    \centering
    \includegraphics[width=\textwidth]{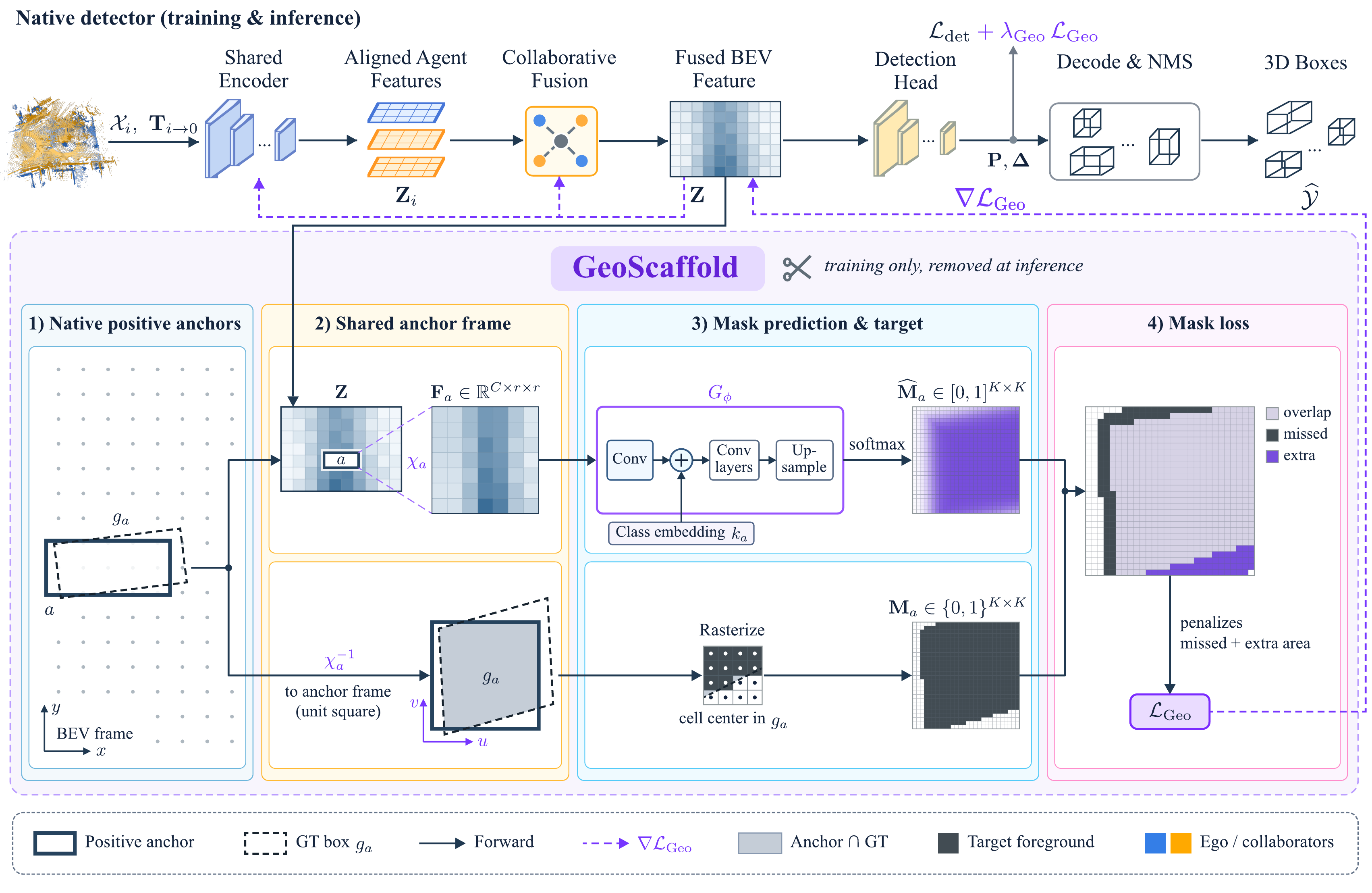}
    \caption{\textbf{Overview of GeoScaffold.} Top: the native detection path (training and
    inference). Bottom: the GeoScaffold branch (training only). (1)~A native positive anchor $a$
    and its assigned box $g_a$ in receiver BEV coordinates $(x,y)$, given by the detector's own
    assignment. (2)~The anchor-coordinate transform $\chi_a$ maps the unit square
    $[-\tfrac{1}{2},\tfrac{1}{2}]^2$ onto the anchor ($u$ along its length, $v$ along its width);
    it samples the fused map $\mathbf{Z}$ into $\mathbf{F}_a$, and its inverse expresses $g_a$
    in this frame, which all $r\times r$ and $K\times K$ maps share. (3)~$G_{\phi}$ predicts the
    per-cell probabilities $\widehat{\mathbf{M}}_a$ (foreground channel shown); the binary
    foreground mask $\mathbf{M}_a$ marks the cells whose centers lie inside $g_a$. (4)~The prediction,
    thresholded at 0.5 for display, overlaid on the target: cross-entropy and Dice both penalize
    the missed and extra cells. Gradients of $\mathcal{L}_{\mathrm{Geo}}$ flow through
    $\mathbf{Z}$ into the fusion and encoder parameters. Feature maps are schematic, and the
    anchor--box pair and the prediction are illustrative.}
    \label{fig:framework}
\end{figure*}

\section{Method}
\label{sec:method}

Sec.~\ref{sec:introduction} traced unrealized latent geometry to score-based selection. When
every positive anchor of an object carries a well-localized box, whichever of their boxes NMS
retains is well localized. The native localization loss supervises each positive anchor only
through a seven-dimensional box code, which leaves implicit which parts of the anchor the object
covers. To train the detector toward a well-localized box at every positive anchor, GeoScaffold adds this spatial
information as an auxiliary target during training: at each positive anchor of the detector's
own assignment, a lightweight head predicts from the fused BEV which cells of a grid over the
anchor lie inside the assigned ground-truth box (Fig.~\ref{fig:framework}). Sec.~\ref{sec:method-native}
reviews the native detector and its anchor assignment, Sec.~\ref{sec:method-target} defines the
foreground mask, Sec.~\ref{sec:method-objective} describes the predictor and the objective, and
Sec.~\ref{sec:method-deploy} covers training and inference.

\subsection{Native Detector and Anchor Assignment}
\label{sec:method-native}

A receiver, indexed as agent $0$, collaborates with $N-1$ connected agents. Agent
$i\in\{0,\ldots,N-1\}$ observes a LiDAR point cloud $\mathcal{X}_i$, and the pose
$\mathbf{T}_{i\rightarrow 0}$ maps its coordinates into the receiver's frame, with
$\mathbf{T}_{0\rightarrow 0}$ the identity. We write $\mathcal{X}=\{\mathcal{X}_i\}_{i=0}^{N-1}$
and $\mathcal{T}=\{\mathbf{T}_{i\rightarrow 0}\}_{i=0}^{N-1}$. We denote the native detector by
$f_{\mathrm{det}}(\mathcal{X},\mathcal{T};\Theta)$, where
$\Theta=(\theta_e,\theta_f,\theta_h)$ collects its encoder, fusion, and head parameters. The detector computes its
detections in four steps, encoding, fusion, prediction, and decoding:
\begin{equation}
\begin{aligned}
    \mathbf{Z}_i
        &= \operatorname{Enc}_{\theta_e}\!\left(\mathcal{X}_i;\mathbf{T}_{i\rightarrow 0}\right),
        \quad i=0,\ldots,N-1, \\
    \mathbf{Z}
        &= \Phi_{\theta_f}\!\left(\{\mathbf{Z}_i\}_{i=0}^{N-1}\right), \\
    (\mathbf{P},\boldsymbol{\delta})
        &= \operatorname{Head}_{\theta_h}(\mathbf{Z}), \\
    \widehat{\mathcal{Y}}
        &= \operatorname{Dec}(\mathbf{P},\boldsymbol{\delta};\mathcal{A}).
\end{aligned}
    \label{eq:native-pipeline}
\end{equation}
The shared encoder $\operatorname{Enc}_{\theta_e}$ produces agent $i$'s BEV feature map
$\mathbf{Z}_i$ in the receiver's frame, applying the pose to the points or to the features,
depending on the method. The fusion operator $\Phi_{\theta_f}$ combines these maps into the
fused BEV map $\mathbf{Z}\in\mathbb{R}^{C\times H_z\times W_z}$. Over the native anchor set $\mathcal{A}$, the detection
head $\operatorname{Head}_{\theta_h}$ predicts classification logits
$\mathbf{P}\in\mathbb{R}^{|\mathcal{A}|\times N_{\mathrm{cls}}}$ for $N_{\mathrm{cls}}$ object
classes and anchor-relative box codes
$\boldsymbol{\delta}\in\mathbb{R}^{|\mathcal{A}|\times 7}$~\cite{SECOND,PointPillars}. The decoder
$\operatorname{Dec}$ decodes the box codes, keeps boxes whose class score exceeds a threshold, and
applies rotated NMS, yielding the detection set $\widehat{\mathcal{Y}}$.

Training uses the detector's native target assignment. Let $\mathcal{Y}$ be the set of annotated
boxes. Anchors are generated per class, and each anchor is labeled positive, negative, or ignored by
thresholding (at 0.6 and 0.45) the IoU of its axis-aligned (stand-up) BEV rectangle with the
boxes of its class; the highest-overlap anchor of each box is also positive. We write the
result as $(\mathcal{A}^{+},\pi,\{k_a\})=\operatorname{Assign}_{\mathrm{det}}(\mathcal{A},\mathcal{Y})$,
where $\mathcal{A}^{+}\subseteq\mathcal{A}$ is the positive-anchor set, $\pi(a)\in\mathcal{Y}$ is
the box assigned to $a\in\mathcal{A}^{+}$, and $k_a\in\{1,\ldots,N_{\mathrm{cls}}\}$ is its class;
we abbreviate $g_a=\pi(a)$. The stand-up IoU only decides which anchors are positive; the foreground
mask of Sec.~\ref{sec:method-target} uses the oriented box $g_a$. One box can receive several positive anchors. They share its
classification target, and their decoded boxes compete in NMS. GeoScaffold reuses this
assignment unchanged, so its supervision reaches every positive anchor that the native losses
train, including each object's top-scored positive anchor, which NMS favors.

\subsection{Anchor-Canonical Foreground Mask}
\label{sec:method-target}

The native box code expresses the box relative to the anchor. GeoScaffold likewise defines its
target in a yaw-aligned and size-normalized anchor frame, so that the
target depends only on the assigned box relative to the anchor. Write a BEV anchor as $a=(\mathbf{c}_a,l_a,w_a,\psi_a)$ with center
$\mathbf{c}_a\in\mathbb{R}^{2}$, length $l_a>0$, width $w_a>0$, and yaw $\psi_a$, and let
$\mathbf{D}_a=\operatorname{diag}(l_a,w_a)$. The anchor-coordinate transform
$\chi_a:\mathbb{R}^{2}\rightarrow\mathbb{R}^{2}$ maps canonical coordinates
$\boldsymbol{\xi}=(u,v)^{\mathsf T}$ to metric BEV coordinates,
\begin{equation}
\begin{aligned}
    \chi_a(\boldsymbol{\xi})
        &= \mathbf{c}_a + \mathbf{R}(\psi_a)\mathbf{D}_a\boldsymbol{\xi}, \\
    \mathbf{R}(\psi)
        &=
        \begin{bmatrix}
            \cos\psi & -\sin\psi \\
            \sin\psi & \cos\psi
        \end{bmatrix},
\end{aligned}
    \label{eq:anchor-chart}
\end{equation}
where $\mathbf{R}(\psi)$ is the counterclockwise rotation by $\psi$, so $u$ runs along the
anchor's length and $v$ along its width. Let $\Omega(\cdot)\subset\mathbb{R}^{2}$ denote the
oriented BEV rectangle of an anchor or box; $\chi_a$ maps the unit square
$\mathcal{Q}=[-\tfrac{1}{2},\tfrac{1}{2}]^2$ onto $\Omega(a)$.

For a grid resolution $m$, the rows $p$ and columns $q$, with $p,q\in\{0,\ldots,m-1\}$, index the
cell centers
$\boldsymbol{\xi}^{(m)}_{pq}=\bigl(\tfrac{q+1/2}{m}-\tfrac{1}{2},\,
\tfrac{p+1/2}{m}-\tfrac{1}{2}\bigr)^{\mathsf T}\in\mathcal{Q}$. At mask resolution $K$, the
foreground mask of $a\in\mathcal{A}^{+}$ is the binary raster with entries, for
$p,q\in\{0,\ldots,K-1\}$,
\begin{equation}
    \mathbf{M}_a[p,q]
        = \mathbf{1}\!\left[
            \chi_a\bigl(\boldsymbol{\xi}^{(K)}_{pq}\bigr)
            \in\operatorname{int}\Omega(g_a)
        \right],
    \label{eq:foreground-mask}
\end{equation}
where $\mathbf{1}[\cdot]$ is the indicator function and $\operatorname{int}$ denotes the set
interior. The $K\times K$ cells tile the anchor, each covering an $l_a/K\times w_a/K$ patch, and
each entry records whether the center of its patch lies inside the assigned box. Thus
$\mathbf{M}_a\in\{0,1\}^{K\times K}$ samples the part of the assigned box inside the anchor,
$\Omega(a)\cap\Omega(g_a)$, and requires no annotation beyond the boxes.

The raster describes, cell by cell, how the assigned object covers the anchor, which the
box code leaves implicit; the relative center offset, box extent, and yaw difference can each
change it. It records only the part of the box inside the anchor at resolution $K$ and does not
determine the box (every box that covers the whole anchor yields an all-ones raster), so the
native regression loss still supervises the complete continuous box, including its vertical
parameters.

\subsection{Auxiliary Predictor and Objective}
\label{sec:method-objective}

The auxiliary input is sampled through the same transform as the target. For
$a\in\mathcal{A}^{+}$, let $\mathbf{Z}$ be the fused map of the sample that contains $a$. At ROI
resolution $r$, the anchor-aligned feature $\mathbf{F}_a\in\mathbb{R}^{C\times r\times r}$ has
entries $\mathbf{F}_a[:,p,q]=\operatorname{Bilinear}\bigl(\mathbf{Z},
\chi_a(\boldsymbol{\xi}^{(r)}_{pq})\bigr)$, where $\operatorname{Bilinear}(\mathbf{Z},\mathbf{x})$
bilinearly interpolates $\mathbf{Z}$ at the metric location $\mathbf{x}$, as in spatial
transformers and aligned region sampling~\cite{Jaderberg2015SpatialTransformer,He_2017_ICCV},
with feature vectors at BEV cell centers and zeros outside the map. $\mathbf{F}_a$ and
$\mathbf{M}_a$ thus cover the same anchor window on cell-centered grids: $\mathbf{F}_a$ is the
input of the auxiliary prediction, and $\mathbf{M}_a$ is its target.

A lightweight predictor $G_{\phi}$ with parameters $\phi$ maps $\mathbf{F}_a$ and the class $k_a$
to background and foreground logits $\mathbf{S}_a=G_{\phi}(\mathbf{F}_a,k_a)$. It projects
$\mathbf{F}_a$ to $C_h$ hidden channels, adds a learned embedding of $k_a$, and applies spatial
convolutions followed by two learned twofold upsampling stages, so its logits have resolution
$4r$; the default $r=7$ gives $K=28$, and the logits are resized bilinearly for other mask resolutions
(App.~\ref{sec:experimental-protocols}). All classes share the spatial predictor, and the class
embedding lets it adapt the prediction to the object category. A per-cell softmax over
$\mathbf{S}_a$ gives the probabilities $\widehat{\mathbf{M}}_a[\kappa,p,q]$, where
$\kappa\in\{0,1\}$ indexes background and foreground.

Let $\tau_{a,pq}=\mathbf{M}_a[p,q]$ be the target and
$\mu_{a,pq}=\widehat{\mathbf{M}}_a[1,p,q]$ the predicted foreground probability, and let
$n_a=\sum_{p,q}\tau_{a,pq}$ count the foreground target cells; every sum ranges over
$p,q\in\{0,\ldots,K-1\}$. With $\epsilon>0$ the floating-point machine epsilon, the per-anchor
mask loss is $\ell_{\mathrm{mask},a}=\ell_{\mathrm{CE},a}+\ell_{\mathrm{Dice},a}$ with
\begin{equation}
\begin{aligned}
    \ell_{\mathrm{CE},a}
        &= -\frac{1}{K^2}\sum_{p,q}
        \log\widehat{\mathbf{M}}_a[\tau_{a,pq},p,q], \\
    \ell_{\mathrm{Dice},a}
        &= \mathbf{1}[n_a>0]
        \left(
            1-
            \frac{2\sum_{p,q}\mu_{a,pq}\tau_{a,pq}+\epsilon}
            {\sum_{p,q}\mu_{a,pq}+n_a+\epsilon}
        \right),
\end{aligned}
    \label{eq:mask-loss}
\end{equation}
where $\widehat{\mathbf{M}}_a[\tau_{a,pq},p,q]$ is the probability of the target channel.
Cross-entropy gives every cell its own gradient, whereas Dice scores the overlap of the predicted
and target regions relative to their areas, so a small covered region counts as much as a large
one; for an empty raster, the indicator removes the Dice term.

For a minibatch $\mathcal{B}$, let $\mathcal{A}^{+}_{\mathcal{B}}$ be its native positive anchors
and $N_{+}=|\mathcal{A}^{+}_{\mathcal{B}}|$. Let $\mathcal{L}_{\mathrm{det}}(\Theta)$ be the
detector's full native objective, including its classification, localization, and
method-specific terms, and let $\lambda_{\mathrm{Geo}}\geq 0$ weight the auxiliary loss ($\lambda_{\mathrm{Geo}}=1$ by default). The
auxiliary loss averages the mask loss over positive anchors, giving each the same weight, and
the training objective is
\begin{equation}
\begin{aligned}
    \mathcal{L}_{\mathrm{Geo}}
        &=
        \begin{cases}
            \displaystyle\frac{1}{N_{+}}
            \sum_{a\in\mathcal{A}^{+}_{\mathcal{B}}}
            \ell_{\mathrm{mask},a}, & N_{+}>0, \\
            0, & N_{+}=0,
        \end{cases} \\
    \mathcal{L}(\Theta,\phi)
        &= \mathcal{L}_{\mathrm{det}}(\Theta)
        + \lambda_{\mathrm{Geo}}\mathcal{L}_{\mathrm{Geo}}(\Theta,\phi).
\end{aligned}
    \label{eq:total-objective}
\end{equation}
$\mathcal{L}_{\mathrm{Geo}}$ involves only positive anchors; negative and ignored anchors are
handled by $\mathcal{L}_{\mathrm{det}}$ alone. The window of $a$ contains the BEV location at
which the head regresses the box code of $a$. Sampling $\mathbf{F}_a$ over this window lets
$\mathcal{L}_{\mathrm{Geo}}$ supervise the fused features from which this box is regressed, and
its gradient reaches the fusion operator and the encoder. The fused BEV is thus trained to encode, at each positive anchor, which parts of the
anchor the object covers.

\subsection{Training and Inference}
\label{sec:method-deploy}

Training minimizes Eq.~\eqref{eq:total-objective} jointly over $\Theta$ and $\phi$ with the optimizer,
learning-rate schedule, and data augmentation of the native detector. At inference, the auxiliary branch is skipped,
and the detector runs the native path with the learned parameters $\Theta^{\star}$,
$f_{\mathrm{deploy}}(\mathcal{X},\mathcal{T})=f_{\mathrm{det}}(\mathcal{X},\mathcal{T};
\Theta^{\star})$. Its inference graph, including communication, fusion, the detection head,
decoding, and NMS, has the parameter count and FLOPs of the native detector, and the predictor
parameters $\phi^{\star}$ are not used; only the values of $\Theta$ differ from those of the Native detector. This training-only use
of an auxiliary branch follows structured-label regularization and removable auxiliary
learning~\cite{Mostajabi_2018_CVPR,He_2020_CVPR,AuxiliaryTraining}.

\section{Experiments}
\label{sec:experiments}

\subsection{Experimental Setup}
\label{sec:experimental-setup}

\begin{table}[t]
    \centering
    \caption{Latent and realized geometry of the Native V2X-ViT detector
    (crossing1/2 validation). AP and mean IoU are in percent. (a) Only box coordinates change; scores,
    classes, count, and order stay fixed. The legal-lineage oracle may replace an output box
    only with a same-class proposal that it suppressed in NMS. The proposal-pool oracle (any
    unused same-class proposal that does not lower IoU) and the GT rebox ceiling (the
    ground-truth box) change only the outputs matched at IoU$\geq$0.5, which also define the
    matched mean IoU. (b) Ego-only and collaborative boxes at each object's
    highest-overlap positive anchor on collaborative frames; mean IoU and object counts.}
    \label{tab:problem-diagnostics}
    \footnotesize
    \setlength{\tabcolsep}{4.0pt}
    \renewcommand{\arraystretch}{1.0}

    \textbf{(a) Fixed-set localization headroom}\par\vspace{2pt}
    \begin{tabular*}{\columnwidth}{@{\extracolsep{\fill}}l rrr@{}}
        \toprule
        \textbf{Geometry readout} & AP@0.5 & AP@0.7 & \makecell{Matched\\mean IoU} \\
        \midrule
        Native readout         & 26.38 & 11.16 & 76.18 \\
        Legal-lineage oracle   & 29.72 & 15.62 & 79.74 \\
        Proposal-pool oracle   & 26.38 & 16.41 & 80.89 \\
        GT rebox ceiling       & 26.38 & 26.38 & 100.00 \\
        \bottomrule
    \end{tabular*}

    \vspace{5pt}
    \textbf{(b) Same-anchor collaborative regression}\par\vspace{2pt}
    \begin{tabular*}{\columnwidth}{@{\extracolsep{\fill}}l rr@{}}
        \toprule
        \textbf{Input at the same anchor} & Ego-only & Collaborative \\
        \midrule
        Mean IoU & 67.81 & 73.17 \\
        Objects with the higher IoU & 24,150 & 51,737 \\
        Objects with IoU$\geq$0.7 only here & 2,986 & 13,553 \\
        \bottomrule
    \end{tabular*}
\end{table}
\begin{table*}[t]
    \centering
    \caption{Results on the UrbanIng-V2X crossing1/2 validation split (AP in percent).
    Ego-only, Early Fusion, and Late Fusion are reference detectors without a $+$Geo
    counterpart. $\Delta$ is $\text{GeoScaffold}-\text{Native}$ in percentage points;
    \textcolor{ForestGreen}{green} and \textcolor{BrickRed}{red} cells mark gains and drops.}
    \label{tab:main-results}
    \footnotesize
    \renewcommand{\arraystretch}{1.0}
    \begin{tabular*}{\textwidth}{@{\extracolsep{\fill}}l rrr rrr rrr@{}}
        \toprule
        \multirow{2}{*}{\textbf{Fusion method}} &
        \multicolumn{3}{c}{\textbf{AP@0.3}} &
        \multicolumn{3}{c}{\textbf{AP@0.5}} &
        \multicolumn{3}{c}{\textbf{AP@0.7}} \\
        \cmidrule(lr){2-4}\cmidrule(lr){5-7}\cmidrule(lr){8-10}
        & Native & $+$Geo & $\Delta$ & Native & $+$Geo & $\Delta$ &
        Native & $+$Geo & $\Delta$ \\
        \midrule
        \multicolumn{10}{@{}l}{\textit{Reference detectors}} \\
        Ego-only & 22.55 & \tabna & \tabna & 16.12 & \tabna & \tabna & 5.41 & \tabna & \tabna \\
        Early Fusion & 23.76 & \tabna & \tabna & 17.35 & \tabna & \tabna & 6.46 & \tabna & \tabna \\
        Late Fusion  & 27.25 & \tabna & \tabna & 19.40 & \tabna & \tabna & 6.88 & \tabna & \tabna \\
        \midrule
        \multicolumn{10}{@{}l}{\textit{Fusion methods with Native/$+$Geo pairs}} \\
        AttFuse~\cite{OPV2V} & 24.25 & 25.19 & \tabgain{+0.94} & 18.18 & 18.89 & \tabgain{+0.71} &  7.16 &  7.30 & \tabgain{+0.15} \\
        F-Cooper~\cite{Fcooper} & 31.62 & 33.12 & \tabgain{+1.49} & 25.35 & 27.43 & \tabgain{+2.08} & 11.19 & 12.57 & \tabgain{+1.38} \\
        CoBEVT~\cite{CoBEVT} & 28.27 & 28.53 & \tabgain{+0.26} & 22.45 & 22.94 & \tabgain{+0.49} &  8.76 &  9.07 & \tabgain{+0.32} \\
        Where2comm~\cite{where2comm} & 30.21 & 30.90 & \tabgain{+0.68} & 24.89 & 26.16 & \tabgain{+1.27} & 11.13 & 11.78 & \tabgain{+0.65} \\
        V2X-ViT~\cite{v2x-vit} & 33.86 & 36.35 & \tabgain{+2.49} & 26.38 & 29.08 & \tabgain{+2.71} & 11.16 & 12.58 & \tabgain{+1.42} \\
        V2VNet~\cite{v2vnet} & 22.84 & 23.48 & \tabgain{+0.64} & 17.11 & 17.95 & \tabgain{+0.84} &  6.65 &  7.07 & \tabgain{+0.43} \\
        DiscoNet~\cite{disconet} & 30.64 & 30.84 & \tabgain{+0.21} & 25.52 & 25.75 & \tabgain{+0.24} & 11.37 & 11.61 & \tabgain{+0.25} \\
        CoAlign~\cite{Coalign} & 28.22 & 28.27 & \tabgain{+0.05} & 19.96 & 20.10 & \tabgain{+0.15} & 7.38 & 7.45 & \tabgain{+0.07} \\
        ERMVP~\cite{ermvp} & 29.19 & 29.86 & \tabgain{+0.67} & 23.76 & 24.15 & \tabgain{+0.39} & 10.04 & 10.72 & \tabgain{+0.68} \\
        HEAL~\cite{HEAL} & 26.59 & 28.12 & \tabgain{+1.52} & 20.72 & 22.53 & \tabgain{+1.81} & 9.13 & 9.41 & \tabgain{+0.29} \\
        S-AdaFusion~\cite{Qiao_2023_WACV} & 29.56 & 29.99 & \tabgain{+0.43} & 21.43 & 21.49 & \tabgain{+0.06} &  7.61 &  7.99 & \tabgain{+0.38} \\
        \bottomrule
    \end{tabular*}

    \vspace{3pt}
    \begin{minipage}{\textwidth}
        \footnotesize
        HEAL uses its homogeneous LiDAR Stage-1 Pyramid Fusion backbone, and S-AdaFusion its
        homogeneous LiDAR intermediate-fusion setting.
    \end{minipage}
\end{table*}
\begin{table*}[t]
    \centering
    \caption{Generalization. On crossing3, a held-out UrbanIng-V2X intersection, the crossing1/2
    final checkpoints are evaluated without retraining or checkpoint selection; V2V4Real pairs
    are trained from scratch. AP is in percent, and
    $\Delta=\text{GeoScaffold}-\text{Native}$ in percentage points.
    App.~\ref{sec:complete-generalization} adds AP@0.3
    and the reference detectors. $^{\dagger}$No detections on V2V4Real in either arm.}
    \label{tab:generalization}
    \footnotesize
    \setlength{\tabcolsep}{2.6pt}
    \renewcommand{\arraystretch}{1.0}
    \begin{tabular*}{\textwidth}{@{\extracolsep{\fill}}l rrr rrr rrr rrr@{}}
        \toprule
        & \multicolumn{6}{c}{\textbf{UrbanIng-V2X crossing3 (held out)}} &
        \multicolumn{6}{c}{\textbf{V2V4Real test}} \\
        \cmidrule(lr){2-7}\cmidrule(lr){8-13}
        & \multicolumn{3}{c}{AP@0.5} & \multicolumn{3}{c}{AP@0.7} &
        \multicolumn{3}{c}{AP@0.5} & \multicolumn{3}{c}{AP@0.7} \\
        \cmidrule(lr){2-4}\cmidrule(lr){5-7}\cmidrule(lr){8-10}\cmidrule(lr){11-13}
        \textbf{Fusion method} & Native & $+$Geo & $\Delta$ & Native & $+$Geo & $\Delta$ & Native & $+$Geo & $\Delta$ & Native & $+$Geo & $\Delta$ \\
        \midrule
        AttFuse & 19.75 & 20.05 & \tabgain{+0.30} & 7.43 & 7.90 & \tabgain{+0.47} & 63.71 & 67.28 & \tabgain{+3.57} & 37.41 & 38.38 & \tabgain{+0.97} \\
        F-Cooper & 22.70 & 29.39 & \tabgain{+6.69} & 9.98 & 13.83 & \tabgain{+3.85} & 63.44 & 66.45 & \tabgain{+3.01} & 36.29 & 38.41 & \tabgain{+2.13} \\
        CoBEVT & 20.70 & 21.52 & \tabgain{+0.83} & 9.20 & 9.49 & \tabgain{+0.29} & 67.12 & 69.08 & \tabgain{+1.96} & 39.39 & 43.76 & \tabgain{+4.36} \\
        Where2comm & 22.48 & 23.41 & \tabgain{+0.93} & 10.65 & 11.08 & \tabgain{+0.43} & 59.47 & 61.29 & \tabgain{+1.83} & 31.68 & 34.50 & \tabgain{+2.83} \\
        V2X-ViT & 21.99 & 22.91 & \tabgain{+0.92} & 9.67 & 10.01 & \tabgain{+0.34} & 67.11 & 67.81 & \tabgain{+0.71} & 36.35 & 39.74 & \tabgain{+3.39} \\
        V2VNet$^{\dagger}$ & 15.37 & 15.76 & \tabgain{+0.39} & 6.01 & 6.12 & \tabgain{+0.11} & \textendash & \textendash & \textendash & \textendash & \textendash & \textendash \\
        DiscoNet & 22.90 & 24.12 & \tabgain{+1.23} & 11.14 & 13.20 & \tabgain{+2.07} & 54.58 & 58.55 & \tabgain{+3.96} & 27.15 & 28.81 & \tabgain{+1.66} \\
        CoAlign & 13.97 & 13.74 & \tabdrop{-0.23} & 4.91 & 4.87 & \tabdrop{-0.03} & 50.47 & 53.22 & \tabgain{+2.74} & 26.68 & 27.94 & \tabgain{+1.27} \\
        ERMVP & 22.67 & 24.45 & \tabgain{+1.78} & 11.08 & 10.37 & \tabdrop{-0.71} & 67.34 & 67.81 & \tabgain{+0.48} & 41.59 & 42.41 & \tabgain{+0.82} \\
        HEAL & 13.54 & 13.90 & \tabgain{+0.36} & 4.48 & 4.90 & \tabgain{+0.43} & 69.89 & 70.54 & \tabgain{+0.65} & 37.19 & 40.13 & \tabgain{+2.94} \\
        S-AdaFusion & 19.86 & 19.58 & \tabdrop{-0.28} & 7.05 & 7.04 & \tabdrop{-0.01} & 61.95 & 62.09 & \tabgain{+0.14} & 31.90 & 33.58 & \tabgain{+1.67} \\
        \bottomrule
    \end{tabular*}
\end{table*}
\begin{table*}[t]
    \centering
    \caption{Design study on the UrbanIng-V2X crossing1/2 validation split (AP and soft IoU in
    percent; \colorbox{tableAccent}{blue}: default setting). \textbf{(a)} All auxiliaries act on
    the same native positive anchors; the box-code auxiliary regresses the native 7-D box code
    with a parameter-matched head. \textbf{(d)} Soft IoU between the foreground mask and its prediction by a
    linear probe of the fused ROI on collaborative validation frames; the prior uses only the
    class and anchor template.}
    \label{tab:geometry-ablation}
    \footnotesize
    \setlength{\tabcolsep}{3.0pt}
    \renewcommand{\arraystretch}{1.0}

    \textbf{(a) Auxiliary representation and objective}\par\vspace{2pt}
    {\setlength{\tabcolsep}{7.8pt}%
    \begin{tabular}{@{}ll rrr rrr rrr@{}}
        \toprule
        & & \multicolumn{3}{c}{\textbf{F-Cooper}} & \multicolumn{3}{c}{\textbf{V2X-ViT}} &
        \multicolumn{3}{c}{\textbf{Where2comm}} \\
        \cmidrule(lr){3-5}\cmidrule(lr){6-8}\cmidrule(lr){9-11}
        \textbf{Auxiliary} & \textbf{Loss} & AP@0.3 & AP@0.5 & AP@0.7 & AP@0.3 & AP@0.5 &
        AP@0.7 & AP@0.3 & AP@0.5 & AP@0.7 \\
        \midrule
        None (Native) & \tabna & 31.62 & 25.35 & 11.19 & 33.86 & 26.38 & 11.16 & 30.21 & 24.89 & 11.13 \\
        7-D box code & SmoothL1 & 32.64 & 26.22 & 11.47 & 35.56 & 27.96 & 11.38 & 32.05 & 25.59 & 11.36 \\
        Foreground mask & CE & 33.17 & 27.38 & 12.22 & 36.12 & 28.76 & 12.05 & 30.56 & 25.59 & 11.84 \\
        Foreground mask & Dice & 34.67 & 28.38 & 12.15 & 33.70 & 26.53 & 11.73 & 30.48 & 25.52 & 11.84 \\
        \rowcolor{tableAccent}
        Foreground mask & CE $+$ Dice & 33.12 & 27.43 & 12.57 & 36.35 & 29.08 & 12.58 & 30.90 & 26.16 & 11.78 \\
        \bottomrule
    \end{tabular}}

    \vspace{4pt}
    \begin{minipage}[t]{0.29\textwidth}
        \centering
        \textbf{(b) Supervised feature (V2X-ViT)}\par\vspace{2pt}
        \begin{adjustbox}{max width=\linewidth}
        \begin{tabular}{@{}l rrr@{}}
            \toprule
            \textbf{Feature} & AP@0.3 & AP@0.5 & AP@0.7 \\
            \midrule
            None (Native) & 33.86 & 26.38 & 11.16 \\
            Ego feature & 35.79 & 28.46 & 12.25 \\
            \rowcolor{tableAccent}
            Fused feature & \textbf{36.35} & \textbf{29.08} & \textbf{12.58} \\
            \bottomrule
        \end{tabular}
        \end{adjustbox}
    \end{minipage}\hfill
    \begin{minipage}[t]{0.29\textwidth}
        \centering
        \textbf{(c) Sensitivity (V2X-ViT)}\par\vspace{2pt}
        \begin{adjustbox}{max width=\linewidth}
        \begin{tabular}{@{}l rrr@{}}
            \toprule
            \textbf{Setting} & AP@0.3 & AP@0.5 & AP@0.7 \\
            \midrule
            $\lambda_{\mathrm{Geo}}=0.5$ & 35.56 & 28.07 & 12.25 \\
            \rowcolor{tableAccent}
            $\lambda_{\mathrm{Geo}}=1$ & \textbf{36.35} & \textbf{29.08} & \textbf{12.58} \\
            $\lambda_{\mathrm{Geo}}=2$ & 34.75 & 28.47 & 11.76 \\
            \midrule
            $K=14$ & 35.19 & 28.27 & 11.63 \\
            \rowcolor{tableAccent}
            $K=28$ & \textbf{36.35} & \textbf{29.08} & \textbf{12.58} \\
            $K=56$ & 35.12 & 28.09 & 11.89 \\
            \bottomrule
        \end{tabular}
        \end{adjustbox}
    \end{minipage}\hfill
    \begin{minipage}[t]{0.37\textwidth}
        \centering
        \textbf{(d) Mask readability}\par\vspace{2pt}
        \begin{adjustbox}{max width=\linewidth}
        \begin{tabular}{@{}l rr@{}}
            \toprule
            & \multicolumn{2}{c}{\textbf{Soft IoU $\uparrow$}} \\
            \cmidrule(l){2-3}
            \textbf{Probe input} & F-Cooper & V2X-ViT \\
            \midrule
            Class $+$ template prior & 59.86 & 59.80 \\
            Native fused ROI & 67.41 & 65.86 \\
            \rowcolor{tableAccent}
            $+$Geo fused ROI & \textbf{71.04} & \textbf{71.51} \\
            $+$Geo ROI, permuted & 62.81 & 57.09 \\
            \bottomrule
        \end{tabular}
        \end{adjustbox}
    \end{minipage}
\end{table*}

We evaluate on two real-world collaborative datasets: UrbanIng-V2X~\cite{urbaningv2x2025}, which
records connected vehicles and roadside infrastructure at three intersections, and
V2V4Real~\cite{v2v4real}, which records two connected vehicles. On UrbanIng-V2X, we train and
validate on crossing1/2 and hold out crossing3 as an unseen intersection. We attach GeoScaffold
to 11 anchor-based fusion methods built on PointPillars~\cite{PointPillars} in OpenCOOD~\cite{OPV2V}
(Table~\ref{tab:main-results}) and write $+$Geo for a detector trained with it. Each
Native/$+$Geo pair shares the architecture and training protocol, and we evaluate the final
checkpoint of each model. We report BEV AP at IoU thresholds of 0.3, 0.5, and 0.7 (AP@0.3,
AP@0.5, and AP@0.7) in percent with each dataset's official evaluator; on UrbanIng-V2X, AP
averages four classes (Vehicle, Heavy Vehicle, 2Wheeler, Pedestrian).
App.~\ref{sec:experimental-protocols} details the splits, metrics, and per-method settings.

\subsection{Latent and Realized Geometry}
\label{sec:exp-realization}

\paragraph{Geometry lost at selection.}
Table~\ref{tab:problem-diagnostics} measures, on the Native V2X-ViT detector, the gap that
motivates GeoScaffold. At the highest-overlap positive anchor of each object, the box decoded from
collaborative input has a higher IoU than the ego-only box for 51,737 objects and a lower IoU
for 24,150; mean IoU rises from 67.81\% to 73.17\%, and 13,553 objects reach IoU $\geq$ 0.7 only
with collaboration. The readout, however, must still choose among the collaborative boxes. Guided
by ground truth, the legal-lineage oracle may replace each output box only with a same-class
proposal that this box suppressed in NMS; with scores, classes, count, and order fixed, it raises AP@0.7 from 11.16 to
15.62. Better-localized boxes thus already exist among the proposals that NMS suppresses.

\paragraph{Localization at fixed scores.}
We keep the Native classification outputs of F-Cooper and V2X-ViT and replace only their box
codes with those of the $+$Geo models, so the anchor scores and their ranking stay fixed. In
Fig.~\ref{fig:teaser-fixed-winner}, the class-balanced fraction of objects whose top-scored positive anchor
yields a box with IoU $\geq$ 0.7 rises from 40.9\% to 45.3\% for F-Cooper and from 36.7\% to
39.4\% for V2X-ViT. AP@0.7 rises from 11.19 to 12.13 and from 11.16 to 11.91,
about two-thirds and half of the gains of the complete
$+$Geo models (Table~\ref{tab:main-results}). The mean IoU over all positive anchors of an
object also rises (App.~\ref{sec:readout-analysis}).

\subsection{Results Across Fusion Methods}
\label{sec:exp-main}

Table~\ref{tab:main-results} compares all 11 pairs on the UrbanIng-V2X validation split.
GeoScaffold improves every method at every threshold, by 0.85, 0.98, and 0.55 points on average
at AP@0.3, AP@0.5, and AP@0.7. The four largest AP@0.5 gains occur for four different fusion
designs: V2X-ViT ($+$2.71), F-Cooper ($+$2.08), HEAL ($+$1.81), and Where2comm ($+$1.27). V2X-ViT and
F-Cooper also gain most at AP@0.7 ($+$1.42 and $+$1.38). At AP@0.7, the four strongest Native
detectors lie between 11.13 and 11.37, and each of their $+$Geo versions exceeds 11.37. For all
11 methods, GeoScaffold also raises the Vehicle AP@0.5 and AP@0.7 and the AP@0.5 at 50--100\,m
(App.~\ref{sec:class-distance-results}).

\subsection{Generalization and Robustness}
\label{sec:exp-generalization}

\paragraph{Held-out intersection.}
On crossing3, an intersection unseen during training, GeoScaffold improves AP@0.5 for nine of
the 11 methods and AP@0.7 for eight (Table~\ref{tab:generalization}).

\paragraph{Second dataset.}
On V2V4Real, we train every pair from scratch on its two-vehicle data
(Table~\ref{tab:generalization}). GeoScaffold improves AP@0.5 and AP@0.7 for all ten methods that
produce detections, by 1.91 and 2.20 points on average; CoBEVT gains the most at AP@0.7
(4.36 points).

\paragraph{Pose error and latency.}
Because GeoScaffold changes the encoder and fusion weights, we test whether five fusion methods
keep their robustness. In App.~\ref{sec:pose-latency}, we add Gaussian noise with standard
deviations of up to 0.6\,m and 0.6$^\circ$ to collaborator poses, or delay collaborator data by up to
500\,ms, at test time. $+$Geo exceeds Native in 87 of the 100 evaluated settings
(five methods, ten levels, AP@0.3 and AP@0.7) and trails it by at most 0.16 points. GeoScaffold thus
preserves their robustness to pose error and latency.

\subsection{Ablation Study}
\label{sec:exp-ablation}

\paragraph{Representation and objective.}
In Table~\ref{tab:geometry-ablation}(a), every auxiliary supervises the same positive anchors.
With CE $+$ Dice, the foreground mask raises AP@0.7 to 12.57 and 12.58 on
F-Cooper and V2X-ViT, whereas a parameter-matched head that regresses the native 7-D box code
reaches 11.47 and 11.38 (Native: 11.19 and 11.16). The foreground mask exceeds the box-code
auxiliary at AP@0.5 and AP@0.7 on all three fusion methods, so its gain surpasses that of further
box-code supervision. Cross-entropy and Dice each improve AP@0.5 and AP@0.7 over
Native on all three methods; their combination gives the best AP@0.7 on F-Cooper and the best AP
at every threshold on V2X-ViT.

\paragraph{Supervised feature and sensitivity.}
On the V2X-ViT validation split, supervising the fused feature outperforms supervising the ego feature before fusion
(12.58 vs.\ 12.25 AP@0.7; Table~\ref{tab:geometry-ablation}(b)). Each tested loss weight
($\lambda_{\mathrm{Geo}}=0.5$, 1, 2) and mask resolution ($K=14$, 28, 56) improves over Native at
all three thresholds (Table~\ref{tab:geometry-ablation}(c)), as does every setting on crossing3
and on AttFuse (App.~\ref{sec:design-controls}).

\paragraph{Mask readability.}
To examine the fused representation itself, we freeze each Native/$+$Geo pair and train one linear
probe on the anchor-aligned features $\mathbf{F}_a$ of both detectors to predict $\mathbf{M}_a$
(Table~\ref{tab:geometry-ablation}(d)). On the collaborative validation frames, GeoScaffold
raises the probe's soft IoU from 67.41\% to 71.04\% for F-Cooper and from 65.86\% to 71.51\% for
V2X-ViT. A feature-free class-and-template prior reaches 59.86\% and 59.80\%, and spatially
permuting the $+$Geo ROI gives 62.81\% and 57.09\%. Both controls fall below the Native ROI:
the probe reads the mask from the spatial layout of the fused BEV, which GeoScaffold makes
more informative (App.~\ref{sec:mask-readability}).
\section{Conclusion}
\label{sec:conclusion}

Collaborative detectors predict better boxes than they output: the geometry that fusion brings to
positive anchors often stays latent when NMS uses Native scores to choose among the boxes at
these anchors.
GeoScaffold addresses this gap during training: an auxiliary head predicts a foreground mask at
each positive anchor, and its loss updates the encoder and fusion modules, training the detector
toward a better-localized box at every positive anchor. The head is removed after training, so the deployed architecture and inference cost stay unchanged. With
Native scores fixed, GeoScaffold improves localization at the top-scored positive anchors of
F-Cooper and V2X-ViT. It also raises AP@0.5 and AP@0.7 for all 11 fusion methods on the
UrbanIng-V2X validation split. Its gains extend to most methods on a held-out intersection and to
all ten methods that produce detections on V2V4Real. Realizing latent geometry in the output
boxes thus deserves the same attention as making it available through fusion. Combining
GeoScaffold with localization-aware scoring is a natural next step.

\section*{LLM Usage Statement}
We used four generative AI tools in this work: Claude Code (Anthropic), Codex (OpenAI), ChatGPT
(OpenAI), and the AI assistant of Overleaf.
\emph{Ideation:} Claude Code, Codex, and ChatGPT took part in discussions of the problem
analysis, candidate designs, and the experimental plan, and they gave reviewer-style feedback on
drafts. The authors decided the research question, the method, the experiments, and the
conclusions.
\emph{Code and experiments:} under the authors' direction, Claude Code and Codex wrote and
edited code for the method, the training and evaluation pipelines, and the analysis and plotting
scripts, and they launched and monitored training and evaluation runs.
\emph{Text, figures, and related work:} Claude Code and Codex drafted and revised text in all
sections, rendered the tables and plots from the evaluation outputs through scripts, and
produced a script-generated layout of Fig.~\ref{fig:framework} that the authors finalized by
hand. They also helped search for related work and check citation metadata. ChatGPT and the
Overleaf assistant were used for translation, language polishing, and \LaTeX{} fixes.
The authors checked the code, results, text, and references, and take full responsibility for
the content of this paper.
{
    \small
    \bibliographystyle{ieeenat_fullname}
    \bibliography{references}
}
\clearpage
\appendix
\setcounter{table}{0}
\renewcommand{\thetable}{A\arabic{table}}
\renewcommand{\theHtable}{appendix.A\arabic{table}}
\setcounter{figure}{0}
\renewcommand{\thefigure}{A\arabic{figure}}
\renewcommand{\theHfigure}{appendix.A\arabic{figure}}
\raggedbottom
\makeatletter
\setlength{\@fptop}{0pt}\setlength{\@fpsep}{12pt}\setlength{\@fpbot}{0pt plus 1fil}
\setlength{\@dblfptop}{0pt}\setlength{\@dblfpsep}{12pt}\setlength{\@dblfpbot}{0pt plus 1fil}
\makeatother
\newpage
\twocolumn[\centering\Large\textbf{\thetitle}\\
\vspace{0.5em}Appendix\\
\vspace{1.0em}]
\noindent\textit{Numbering.} The sections of this appendix are lettered A--D, and its tables and
figures are numbered A1, A2, and so on. Tables, figures, sections, and equations cited without
a letter belong to the main paper.

\section{Experimental Protocols}
\label{sec:experimental-protocols}

\subsection{Data and Evaluation}
\paragraph{Metrics.}
We report BEV average precision at IoU thresholds of 0.3, 0.5, and 0.7 in percent with each
dataset's official evaluator; AP@0.7 is the most sensitive to localization. UrbanIng-V2X AP is the
mean of the four class APs within 100\,m of the receiver, and V2V4Real reports vehicle AP.
$\Delta$ denotes GeoScaffold minus Native in percentage points, computed before rounding. On both datasets, every detector
keeps boxes with scores above 0.2 and applies rotated NMS with an IoU threshold of 0.15.

\paragraph{UrbanIng-V2X.}
The crossing1/2 training and validation splits contain 20 and seven scenarios, respectively;
crossing3 contributes seven held-out scenarios. Their frame counts are 4,000, 1,400, and
1,400. The category mapping groups Car and Van as Vehicle; Truck, Bus, Trailer, and
OtherVehicle as Heavy Vehicle; Cyclist, Motorcycle, and E-Scooter as 2Wheeler; and Pedestrian
and OtherPedestrian as Pedestrian. Distance-resolved results use the evaluator's 0--30, 30--50,
and 50--100\,m intervals. Crossing3 is excluded from training and checkpoint selection.

\paragraph{V2V4Real.}
The training, validation, and test splits contain 32, three, and nine scenes, respectively.
We use synchronized timestamps with observations from both vehicles, yielding 7,105 training,
748 validation, and 1,993 test frames. Vehicle AP follows the dataset's rectangular evaluation
extent, $x\in[-100,100]$ and $y\in[-40,40]$ in meters around the receiver, in place of
UrbanIng-V2X's 100\,m radius. Each model learns from V2V4Real
training data independently, including its teachers or prerequisite detectors where applicable.
On both datasets, Ego-only receives only the receiver's LiDAR, while evaluation retains the
ground-truth annotations from the corresponding collaborative scene.

\paragraph{Spatial discretization.}
Coordinates and voxel sizes below are in meters, with $x$ and $y$ spanning the horizontal plane.
Most UrbanIng-V2X encoders use $x\in[-102.4,102.4]$, $y\in[-40,40]$, and $z\in[-5,5]$,
with voxel size $(0.4,0.4,10)$; V2X-ViT uses $y\in[-38.4,38.4]$.
Early Fusion and Late Fusion use $z\in[-10,10]$ and voxel height 20, and Late Fusion also uses
$x\in[-70.4,70.4]$.
Most V2V4Real encoders use $x\in[-140.8,140.8]$, $y\in[-38.4,38.4]$, and $z\in[-5,3]$,
with voxel size $(0.4,0.4,8)$, including AttFuse.
CoBEVT uses $x\in[-102.4,102.4]$ with the same $y$, $z$, and voxel settings.
Ego-only and Late Fusion use $x\in[-70.4,70.4]$, $y\in[-40,40]$; Early Fusion uses
$x\in[-140.8,140.8]$ and $y\in[-40,40]$. Native/$+$Geo pairs always retain identical spatial settings.
The reference detectors retain their adopted input ranges.

\subsection{Training Recipes and Paired Controls}
\paragraph{Initialization and optimization.}
All detectors use PointPillars encoders in OpenCOOD and train on a single NVIDIA RTX 4090 GPU.
Each $+$Geo model keeps the encoder, fusion module, detection heads, decoder, post-processing,
and method-specific components, such as distillation teachers and pose correction, of its
Native counterpart. Ego-only, Early Fusion, and Late Fusion serve as reference detectors.
Table~\ref{tab:protocols}(a,b) gives the training horizons, auxiliary intervals, optimizer schedules,
and shared native prefixes of all pairs. All pairs use seed 1669408686 and the same data
augmentation and sample order. Training augmentation comprises random flipping along the $x$
axis, global rotation within $[-\pi/4,\pi/4]$, and uniform scaling within $[0.95,1.05]$ on both
datasets. The data-order generator is independent of auxiliary-head initialization. When a pair
shares a native prefix, both arms continue from the same checkpoint and only the $+$Geo arm adds
the auxiliary; both arms initialize the optimizer afresh, and learning-rate schedules keep the
global epoch index. On V2V4Real, both arms use their full dataset-specific schedule from
scratch, with auxiliary supervision from epoch 1.

\begin{table*}[t]
    \centering
    \caption{Training protocols. \textbf{(a) UrbanIng-V2X.} Each row describes the shared Native/$+$Geo recipe; the Geo interval gives the epochs with auxiliary supervision. Reference detectors have no Geo branch. Adam learning rates are shown below; all rows use weight decay $10^{-4}$.}
    \label{tab:protocols}
    \footnotesize
    \setlength{\tabcolsep}{3pt}
    \renewcommand{\arraystretch}{1.24}
    \begin{tabular}{@{}p{0.135\textwidth} >{\centering\arraybackslash}p{0.07\textwidth} >{\centering\arraybackslash}p{0.09\textwidth} p{0.13\textwidth} p{0.20\textwidth} p{0.31\textwidth}@{}}
        \toprule
        \textbf{Method} & \textbf{Epochs} & \textbf{Geo epochs} & \textbf{Optimizer / LR} & \textbf{LR schedule} & \textbf{Method-specific setup} \\
        \midrule
        Ego-only & 20 & \tabna & Adam $1\!\times\!10^{-3}$ & Cosine; warm-up 10 & Receiver LiDAR only \\
        Early Fusion & 20 & \tabna & Adam $1\!\times\!10^{-3}$ & Cosine; warm-up 10 & Aligned raw-point aggregation \\
        Late Fusion & 20 & \tabna & Adam $1\!\times\!10^{-3}$ & Cosine; warm-up 10 & Independent detection; box fusion \\
        \midrule
        AttFuse & 20 & 12--20 & Adam $1\!\times\!10^{-3}$ & Cosine; warm-up 10 & Per-cell agent attention \\
        F-Cooper & 60 & 1--60 & Adam $1\!\times\!10^{-3}$ & Cosine; warm-up 10 & Source-wise max pooling \\
        CoBEVT & 20 & 12--20 & Adam $1\!\times\!10^{-3}$ & Cosine; warm-up 10 & FAX; depth 3, window 4 \\
        Where2comm$^{\dagger}$ & 60 & 12--60 & Adam $1\!\times\!10^{-3}$ & Cosine; warm-up 10 & Multiscale; communication threshold 0.01 \\
        V2X-ViT$^{\dagger}$ & 30 & 12--30 & Adam $1\!\times\!10^{-3}$ & Cosine; warm-up 10 & Transformer depth 1; windows 2/4/8 \\
        V2VNet & 20 & 12--20 & Adam $1\!\times\!10^{-3}$ & Cosine; warm-up 10 & ConvGRU; 1 message-passing iteration; vehicle agents only \\
        DiscoNet$^{\dagger}$ & 40 & 12--40 & Adam $1\!\times\!10^{-3}$ & Cosine; warm-up 10 & Early-Fusion teacher (20 ep.); KD weight 0.5 \\
        CoAlign & 30 & 22--30 & Adam $2\!\times\!10^{-3}$ & Steps 10,15; $\gamma=0.1$ & Local-feature warping; uncertainty pretraining (20 ep.) \\
        ERMVP$^{\dagger}$ & 30 & 12--30 & Adam $1\!\times\!10^{-3}$ & Cosine; warm-up 10 & Sparse fusion; depth 1; ego-loss weight 0.5 \\
        HEAL & 30 & 12--30 & Adam $2\!\times\!10^{-3}$ & Steps 15,25; $\gamma=0.1$ & Homogeneous LiDAR Pyramid Fusion \\
        S-AdaFusion & 30 & 12--30 & Adam $5\!\times\!10^{-4}$ & Steps 10,15; $\gamma=0.2$ & Source max/mean statistics; 3D convolution \\
        \bottomrule
    \end{tabular}
    \vspace{3pt}
    \begin{minipage}{\textwidth}
        \footnotesize
        F-Cooper uses the dataset authors' 60-epoch recipe from scratch in both arms, without optimizer resets. CoAlign shares a 21-epoch native prefix and branches with fresh optimizers for epochs 22--30; the prefix optimizer is reset at epoch 12. The remaining paired methods share an 11-epoch native prefix and use fresh optimizers for epochs 12 onward, with schedules indexed by the global epoch. Cosine schedules warm up from $2\times10^{-4}$ and decay to $2\times10^{-5}$. $^{\dagger}$The shared prefix uses the 20-epoch cosine horizon; the continuation uses the indicated extended horizon. CoAlign's 20-epoch uncertainty pretraining is additional to the fusion-training epochs shown here.
    \end{minipage}
\end{table*}

\begin{table*}[t]
    \centering
    \ContinuedFloat
    \caption{Training protocols (continued). \textbf{(b) V2V4Real.} Each row describes the shared Native/$+$Geo recipe; the Geo interval gives the epochs with auxiliary supervision. Reference detectors have no Geo branch. Adam learning rates are shown below; all rows use weight decay $10^{-4}$.}
    \label{tab:protocols-v2v4real}
    \footnotesize
    \setlength{\tabcolsep}{3pt}
    \renewcommand{\arraystretch}{1.24}
    \begin{tabular}{@{}p{0.135\textwidth} >{\centering\arraybackslash}p{0.07\textwidth} >{\centering\arraybackslash}p{0.09\textwidth} p{0.13\textwidth} p{0.20\textwidth} p{0.31\textwidth}@{}}
        \toprule
        \textbf{Method} & \textbf{Epochs} & \textbf{Geo epochs} & \textbf{Optimizer / LR} & \textbf{LR schedule} & \textbf{Method-specific setup} \\
        \midrule
        Ego-only & 60 & \tabna & Adam $2\!\times\!10^{-3}$ & Cosine; warm-up 10 & Receiver LiDAR only \\
        Early Fusion & 60 & \tabna & Adam $2\!\times\!10^{-3}$ & Cosine; warm-up 10 & Aligned raw-point aggregation \\
        Late Fusion & 60 & \tabna & Adam $2\!\times\!10^{-3}$ & Cosine; warm-up 10 & Independent detection; box fusion \\
        \midrule
        AttFuse & 60 & 1--60 & Adam $1\!\times\!10^{-3}$ & Cosine; warm-up 10 & Per-cell attention; compression 32 \\
        F-Cooper & 60 & 1--60 & Adam $2\!\times\!10^{-3}$ & Cosine; warm-up 10 & Source-wise max pooling \\
        CoBEVT & 60 & 1--60 & Adam $1\!\times\!10^{-3}$ & Cosine; warm-up 10 & FAX; depth 3, window 4; compression 32 \\
        Where2comm & 30 & 1--30 & Adam $2\!\times\!10^{-3}$ & Steps 10,20; $\gamma=0.1$ & Multiscale; communication threshold 0.01 \\
        V2X-ViT & 60 & 1--60 & Adam $1\!\times\!10^{-3}$ & Cosine; warm-up 10 & Transformer depth 1; windows 2/4/8 \\
        V2VNet & 60 & 1--60 & Adam $2\!\times\!10^{-3}$ & Cosine; warm-up 10 & ConvGRU; 1 message-passing iteration \\
        DiscoNet & 40 & 1--40 & Adam $2\!\times\!10^{-3}$ & Steps 10,15,40; $\gamma=0.1$ & Early-Fusion teacher (60 ep.); KD weight 0.5 \\
        CoAlign & 30 & 1--30 & Adam $2\!\times\!10^{-3}$ & Steps 10,15; $\gamma=0.1$ & Local-feature warping; uncertainty pretraining (30 ep.) \\
        ERMVP & 60 & 1--60 & Adam $1\!\times\!10^{-3}$ & Cosine; warm-up 10 & Sparse fusion; depth 3; ego-loss weight 0.5 \\
        HEAL & 30 & 1--30 & Adam $2\!\times\!10^{-3}$ & Steps 15,25; $\gamma=0.1$ & LiDAR Pyramid Fusion; $\lambda_{\mathrm{Geo}}=0.5$ \\
        S-AdaFusion & 30 & 1--30 & Adam $5\!\times\!10^{-4}$ & Steps 10,15; $\gamma=0.2$ & Source max/mean statistics; 3D convolution \\
        \bottomrule
    \end{tabular}
    \vspace{3pt}
    \begin{minipage}{\textwidth}
        \footnotesize
        Cosine schedules warm up from $2\times10^{-4}$ and decay to $2\times10^{-5}$. Step lists reproduce the adopted configurations, including milestones at or beyond the last training epoch. Both arms start from scratch; CoAlign's preceding uncertainty detector is trained for 30 epochs with Adam ($10^{-3}$), decays at 10/20, and $\gamma=0.1$.
    \end{minipage}
\end{table*}

\paragraph{Recipe provenance.}
We adapt the public PointPillars implementation of each method to each dataset's geometry and
label space and keep the training horizon of each architecture. We retain
the released V2V4Real recipes where available. Where2comm's V2V4Real run adapts its public
30-epoch DAIR-V2X recipe; DiscoNet uses the 40-epoch PointPillars reproduction maintained in
CoAlign. HEAL and S-AdaFusion use their public homogeneous training recipes with the dataset
interface adapted accordingly; on UrbanIng-V2X, HEAL uses the PointPillars detection objective
described below. Training conditions are matched within each Native/$+$Geo pair.
On V2V4Real, AttFuse uses its released per-cell agent-attention architecture, with a
384-to-256-channel feature reduction and compression factor 32 before fusion.
On both datasets, CoAlign uses a residual multiscale backbone, warps agent-local features into the receiver
frame, and retains the direction-classification head and its native detection objective.
For CoAlign and HEAL, global augmentation conjugates inter-agent transforms so that points
and warped features share one coordinate frame.

\paragraph{Method-specific dependencies.}
DiscoNet retains feature distillation with loss weight 0.5 from a frozen Early-Fusion teacher:
the 20-epoch UrbanIng-V2X teacher or the independently trained 60-epoch V2V4Real teacher.
CoAlign retains uncertainty-aware detector pretraining, per-agent box precomputation, and box
alignment. On UrbanIng-V2X, alignment solves the official agent--landmark pose graph
(at most 1,000 iterations, yaw-variance threshold 0.2); on V2V4Real, it fits a weighted rigid
SE(2) transform (at least two landmarks, 20 iterations). Its
UrbanIng-V2X uncertainty detector is initialized from the native AttFuse
checkpoint and trained for 20 additional epochs; its V2V4Real uncertainty detector is trained
from scratch for 30 epochs. These prerequisite epochs are additional to fusion training.
ERMVP retains sparse fusion/calibration and the additional ego detection loss, with weight 0.5.
HEAL retains single-modality LiDAR Pyramid Fusion and multiscale occupancy supervision, with
occupancy loss weights of 0.4, 0.2, and 0.1 across the three scales. On UrbanIng-V2X, its
detection objective has no direction head and weights classification and regression by 1 and 2.
On V2V4Real, it retains two-bin direction classification, weights positive anchors by 2, and
uses classification, regression, and direction loss weights of 1, 2, and 0.2, respectively.
On UrbanIng-V2X, V2VNet exchanges messages among vehicle agents only; infrastructure agents are excluded.
Where2comm retains its communication mask and adopted operating point, and each arm transmits
the features that this mask selects.

\paragraph{Auxiliary supervision and checkpoint policy.}
GeoScaffold constructs quantized anchor-relative foreground masks from native positive-anchor
assignments and class labels (Sec.~\ref{sec:method}). It uses native fused features
(256 channels) without altering detection targets. The auxiliary uses $7\times7$ ROIs,
128 hidden channels, $K=28$, equally weighted cross-entropy and Dice losses, and a default
weight of $\lambda_{\mathrm{Geo}}=1$ (0.5 for HEAL on V2V4Real). For the resolution study
(Table~\ref{tab:geometry-ablation}(c)), the predictor is unchanged and its $28\times28$ logits are
bilinearly resized to $K=14$ or $K=56$.
The auxiliary is skipped at inference, and each arm is evaluated at the final checkpoint of its
declared horizon.
\section{Geometry Realization Analysis}
\label{sec:geometry-analysis}

\subsection{Collaborative Correction and Candidate Selection}
\label{sec:collaboration-selection}

We analyze the Native F-Cooper and V2X-ViT detectors on the 1,400-frame crossing1/2 validation
split. Coll denotes the collaborative input, and No Coll denotes the ego-only input, which keeps
the same ego voxels and removes the remote inputs; the model weights, ego coordinate frame, box decoder, and evaluation
inputs are unchanged. The statistics below use the 999 collaborative frames, in which the receiver has at least one
collaborator. Positive anchors
and their objects follow the official anchor assignment, independently of predicted scores and
box quality. Each statistic is computed per class, and the four class values are averaged with
equal weight. Table~\ref{tab:selection-statistics} lists the shares that the main paper
cites, with their populations and the figure panels that plot them.

\begin{table*}[t]
    \centering
    \caption{Collaboration and selection statistics of the Native detectors.
    Class-balanced shares (\%) on the 999 collaborative crossing1/2 validation frames. The last
    column names the figure panel that plots each share.}
    \label{tab:selection-statistics}
    \footnotesize
    \setlength{\tabcolsep}{4pt}
    \begin{tabular}{@{}l l rr l@{}}
        \toprule
        \textbf{Statistic} & \textbf{Population (F-Cooper / V2X-ViT)} & \textbf{F-Cooper} &
        \textbf{V2X-ViT} & \textbf{Plotted in} \\
        \midrule
        Coll box has the higher IoU & 165,392 / 165,165 positive anchors &
        67.52 & 66.40 & Fig.~\ref{fig:teaser-nms-population} \\
        NMS keeps the better-localized box & 68,661 / 65,160 same-object NMS pairs &
        48.46 & 46.06 & Figs.~\ref{fig:teaser-nms-population}, \ref{fig:evidence-chain}(b) \\
        Higher-scored anchor yields the higher-IoU box & 137,004 / 136,734 positive-anchor pairs &
        45.31 & 46.78 & Fig.~\ref{fig:teaser-nms-population} \\
        Best-localized box reaches IoU $\geq$ 0.7 & 75,939 / 75,891 objects &
        48.30 & 44.14 & \textendash \\
        Box at top-scored positive anchor reaches IoU $\geq$ 0.7 & same objects &
        40.93 & 36.70 & Figs.~\ref{fig:teaser-fixed-winner}, \ref{fig:evidence-chain}(c) \\
        \quad with GeoScaffold boxes & same objects &
        45.30 & 39.40 & Figs.~\ref{fig:teaser-fixed-winner}, \ref{fig:evidence-chain}(c) \\
        \bottomrule
    \end{tabular}
\end{table*}

\begin{figure*}[t]
    \centering
    \includegraphics[width=\textwidth]{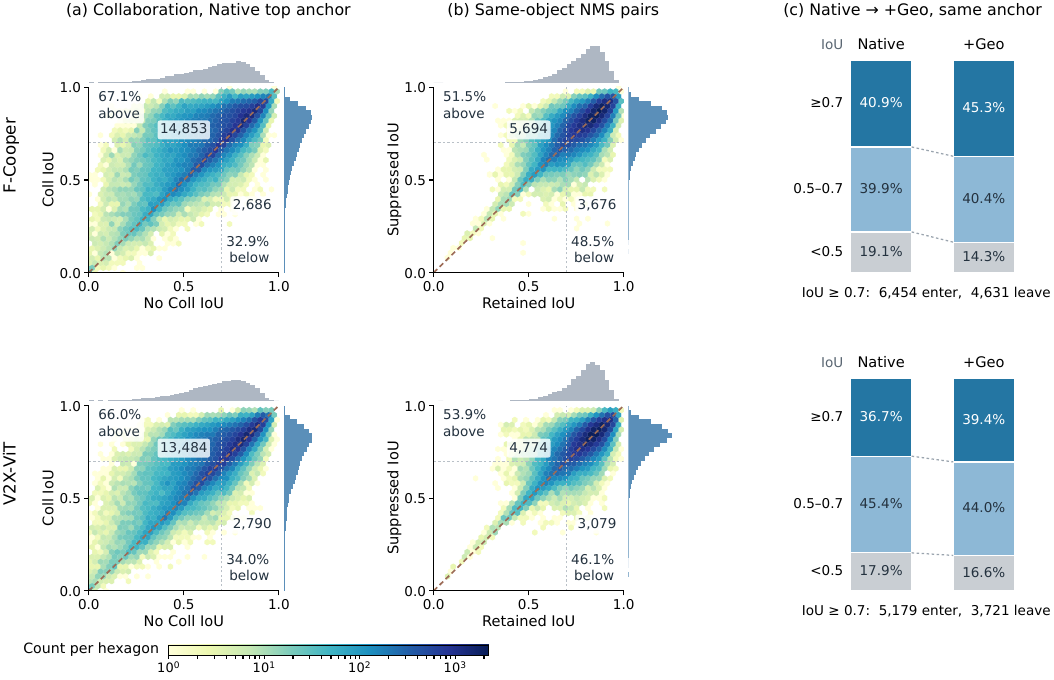}
    \caption{\textbf{Collaborative correction, suppression, and GeoScaffold gains.}
    Each row uses the 999 collaborative validation frames of one detector. Hexagons and marginal
    histograms use all objects or pairs without class weighting; printed percentages and bar
    segments weight the four classes equally, as in Table~\ref{tab:selection-statistics}.
    \textbf{(a)} No Coll and Coll IoU at the Native top-scored positive anchor of every object
    with a positive anchor (75,939 F-Cooper and 75,891 V2X-ViT objects).
    \textbf{(b)} Coll IoU of the retained and the suppressed box in each same-object NMS pair
    (68,661 and 65,160 pairs). In (a) and (b), percentages are shares above or below the dashed
    equal-IoU diagonal, dotted lines mark IoU~0.7, and the two counts give the objects or pairs
    in which only the box on the vertical axis (upper left) or only the box on the horizontal
    axis (lower right) reaches it. Marginal histograms share one density
    scale, and all density panels share one logarithmic color scale.
    \textbf{(c)} Shares of the Native and $+$Geo boxes at the anchors of (a) in three IoU bins;
    the line below each pair of bars counts the objects that enter or leave the IoU $\geq$ 0.7
    bin.}
    \label{fig:evidence-chain}
\end{figure*}

\paragraph{Collaboration corrects boxes at fixed anchors.}
At each positive anchor, we decode one box from the No Coll input and one from the Coll input.
Over 165,392 F-Cooper and 165,165 V2X-ViT positive anchors, the Coll box has the higher IoU at
67.5\% and 66.4\% of anchors. At the Native top-scored positive anchor of each object, the Coll box has
the higher IoU for 67.1\% and 66.0\% of objects. Only the Coll box reaches IoU~0.7 for 14,853
and 13,484 objects, and only the No Coll box does for 2,686 and 2,790
(Fig.~\ref{fig:evidence-chain}(a)).

\paragraph{NMS often suppresses the better-localized box.}
We trace the detector's score threshold, class-agnostic NMS, and post-NMS filtering, and record
the first suppressor of every suppressed candidate. A same-object pair requires both candidates
to belong to the same positive-anchor object with a matching predicted class, the suppressor to
survive the output filtering, and both boxes to lie within the evaluation range; candidates
removed by the top-1,000 score cap are not counted as suppressed. Over 68,661 and 65,160 such
pairs, the retained box has the higher Coll IoU in 48.5\% and 46.1\% of pairs. Only the
suppressed box reaches IoU~0.7 in 5,694 and 4,774 pairs, and only the retained box does in 3,676
and 3,079 pairs (Fig.~\ref{fig:evidence-chain}(b)).

\paragraph{Top-scored and best-localized boxes.}
Of the objects with at least one positive anchor, including anchors below the score threshold
(75,939 and 75,891 objects), the best-localized box across an object's positive anchors has IoU $\geq$ 0.7
for 48.3\% and 44.1\% of objects, and the box at the top-scored positive anchor does so for 40.9\% and
36.7\% (Table~\ref{tab:selection-statistics}; Figs.~\ref{fig:teaser-fixed-winner}
and~\ref{fig:evidence-chain}(c)). The 162 and 161 objects without a positive anchor are excluded
from this statistic.

\paragraph{Scores do not rank localization.}
Within each object, we compare every pair of positive anchors that differ in both score and IoU.
The higher-scored anchor yields the box with the higher IoU in 45.31\% of F-Cooper and 46.78\% of V2X-ViT pairs.

\paragraph{Illustrative case.}
The F-Cooper object in Fig.~\ref{fig:teaser} is selected by a fixed rule: its top-scored
positive anchor suppresses a same-object box with IoU $\geq$ 0.8 and a collaborative IoU gain of
at least 0.08, and qualifying objects are ranked by the GeoScaffold gain at the kept anchor.
Figs.~\ref{fig:teaser-observations}, \ref{fig:teaser-readout}, and~\ref{fig:teaser-geo-case}
label the score and IoU of each box: the retained candidate outscores the suppressed one yet
localizes the object worse, and GeoScaffold's box at the retained anchor nearly matches the
suppressed box.

\paragraph{GeoScaffold at the same anchor.}
Fig.~\ref{fig:evidence-chain}(c) keeps the Native top-scored positive anchor of each object and replaces
the Native box with GeoScaffold's. The box rises to IoU $\geq$ 0.7 for 6,454 F-Cooper and 5,179
V2X-ViT objects and falls below it for 4,631 and 3,721. The class-balanced share of objects at
IoU $\geq$ 0.7 rises from 40.9\% to 45.3\% for F-Cooper and from 36.7\% to 39.4\% for V2X-ViT. App.~\ref{sec:readout-analysis} reports
the corresponding mean IoU and AP.

\subsection{Box Geometry and Score Selection}
\label{sec:readout-analysis}

\paragraph{Protocol.}
On the same 1,400 frames, we combine the dense regression and classification outputs of the
Native and $+$Geo checkpoints before the unchanged native post-processing. Native uses both
Native outputs; Geometry only replaces the regression outputs; Scores only replaces the
classification outputs; $+$Geo replaces both. Replacing the classification outputs changes
confidence, class assignment, filtering, and NMS order. All conditions use the same
input-sampling seed for each frame and the official evaluator, and the Native and $+$Geo conditions reproduce the AP of
Table~\ref{tab:main-results}. The fixed-set statistics use the positive anchors, objects, and
equal class weighting of App.~\ref{sec:collaboration-selection}.

\paragraph{Localization at positive anchors.}
Mean IoU at the top-scored positive anchor is 64.43\% for Native, 66.34\% for Geometry only,
64.56\% for Scores only, and 66.35\% for $+$Geo on F-Cooper, and 63.45\%, 64.32\%,
63.52\%, and 64.35\% on V2X-ViT. Geometry only keeps the Native top-scored positive anchor, so its gain
comes entirely from the box at that anchor. The best-localized box across an object's positive anchors also improves: its
mean IoU rises from 67.03\% with the Native boxes to 68.67\% with GeoScaffold's boxes for
F-Cooper, and from 65.91\% to 66.78\% for V2X-ViT. Averaged over all positive anchors of an object, the mean IoU rises from
64.69\% to 66.51\% for F-Cooper and from 63.69\% to 64.57\% for V2X-ViT.

\begin{figure}[t]
    \centering
    {\footnotesize
    \textcolor[HTML]{AEB7C3}{\rule{0.9em}{0.6em}}\,Native\enspace
    \textcolor[HTML]{5B8FB9}{\rule{0.9em}{0.6em}}\,Geometry only\enspace
    \textcolor[HTML]{D39A52}{\rule{0.9em}{0.6em}}\,Scores only\enspace
    \textcolor[HTML]{55A996}{\rule{0.9em}{0.6em}}\,$+$Geo\par}
    \vspace{0.3em}
    \includegraphics[width=0.74\linewidth]{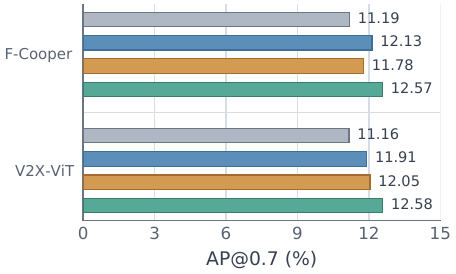}
    \caption{\textbf{Official AP@0.7 after output substitution.}
    Geometry only uses the $+$Geo regression outputs with the Native classification outputs; Scores only
    performs the complementary substitution. Both use the unchanged native decoding and
    post-processing.}
    \label{fig:readout-interventions}
\end{figure}

\paragraph{Final AP.}
For F-Cooper, AP@0.7 is 11.19 for Native, 12.13 for Geometry only, 11.78 for Scores only, and
12.57 for $+$Geo; for V2X-ViT, it is 11.16, 11.91, 12.05, and 12.58
(Fig.~\ref{fig:readout-interventions}). Each substitution improves AP@0.7, and replacing both
outputs gives the highest value. With the $+$Geo scores,
the score--IoU ordering agreement stays near chance, at 47.98\% for F-Cooper and 46.08\% for
V2X-ViT. Geometry only keeps the Native classification outputs and the native post-processing,
and it raises AP@0.7 by 0.94 and 0.74 points. Better boxes alone thus raise AP under the native
selection procedure.

\begin{figure*}[t]
    \centering
    \begin{subfigure}[b]{0.655\textwidth}
        \centering
        \includegraphics[height=1.95in]{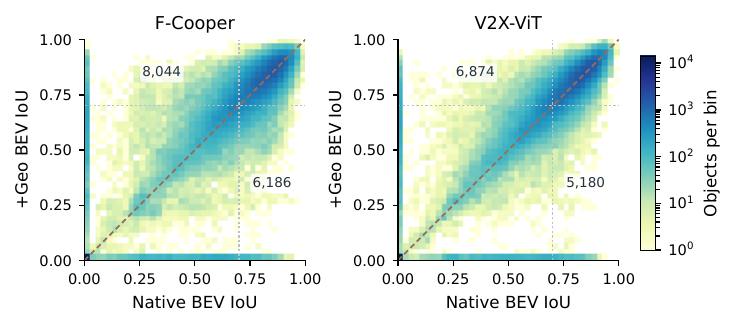}
        \caption{Paired target-level localization quality.}
    \end{subfigure}\hfill
    \begin{subfigure}[b]{0.335\textwidth}
        \centering
        \includegraphics[height=1.95in]{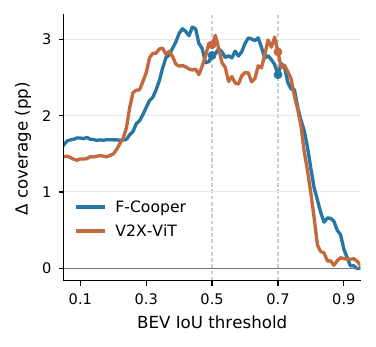}
        \caption{Coverage gain across IoU thresholds.}
    \end{subfigure}
    \caption{\textbf{Output geometry on crossing1/2.}
    Both detectors use all 1,400 validation frames. Predictions are matched one-to-one to
    same-class ground-truth objects by maximum total BEV IoU, separately for each model;
    unmatched objects have IoU zero.
    \textbf{(a)} Paired IoUs of each object. The histograms pool the four classes, use bins of
    width 0.025 and a common logarithmic count scale; the dashed diagonal denotes equal IoU.
    IoU~0.7 is reached only with GeoScaffold by 8,044 F-Cooper and 6,874 V2X-ViT objects, and
    only with Native by 6,186 and 5,180.
    \textbf{(b)} GeoScaffold minus Native coverage, the fraction of objects whose IoU exceeds
    each threshold, averaged equally over the four classes; dots mark IoU~0.5 and~0.7.}
    \label{fig:localization-distributions}
\end{figure*}

\paragraph{Output geometry.}
Fig.~\ref{fig:localization-distributions} compares the final outputs of each model, with its
own scores and NMS, after one-to-one matching to same-class objects. At IoU~0.7, GeoScaffold
raises class-balanced coverage by 2.53 percentage points for F-Cooper and 2.83 for V2X-ViT.

\section{Additional Results}
\label{sec:additional-results}

\subsection{Per-Class and Distance-Resolved Results}
\label{sec:class-distance-results}
Tables~\ref{tab:per-class-analysis} and~\ref{tab:distance-range-analysis} resolve
Table~\ref{tab:main-results} by category and by the evaluator's range bins, using the same final
checkpoints and evaluator.

\begin{table*}[p]
    \centering
    \caption{Per-class analysis on the UrbanIng-V2X crossing1/2 validation split. Each panel reports
    Native, $+$Geo, and their signed percentage-point difference for all three official IoU
    thresholds. \textcolor{ForestGreen}{Green} and \textcolor{BrickRed}{red} $\Delta$ cells mark
    gains and drops. Values come from the same evaluations as
    Table~\ref{tab:main-results}.}
    \label{tab:per-class-analysis}
    \scriptsize
    \setlength{\tabcolsep}{2.8pt}
    \renewcommand{\arraystretch}{0.98}
    \textbf{(a) Vehicle}\par\vspace{1pt}
    \begin{adjustbox}{max width=\textwidth}
    \begin{tabular*}{\textwidth}{@{\extracolsep{\fill}}l rrr rrr rrr@{}}
        \toprule
        \multirow{2}{*}{\textbf{Fusion method}} & \multicolumn{3}{c}{\textbf{AP@0.3}} &
        \multicolumn{3}{c}{\textbf{AP@0.5}} & \multicolumn{3}{c}{\textbf{AP@0.7}} \\
        \cmidrule(lr){2-4}\cmidrule(lr){5-7}\cmidrule(lr){8-10}
        & Native & $+$Geo & $\Delta$ & Native & $+$Geo & $\Delta$ & Native & $+$Geo & $\Delta$ \\
        \midrule
        AttFuse & 58.21 & 59.11 & \tabgain{+0.90} & 48.65 & 49.61 & \tabgain{+0.97} & 24.61 & 25.71 & \tabgain{+1.10} \\
        F-Cooper & 70.34 & 71.90 & \tabgain{+1.56} & 61.19 & 64.12 & \tabgain{+2.92} & 35.49 & 38.50 & \tabgain{+3.01} \\
        CoBEVT & 64.64 & 64.98 & \tabgain{+0.34} & 55.29 & 56.25 & \tabgain{+0.96} & 30.36 & 31.31 & \tabgain{+0.94} \\
        Where2comm & 69.06 & 69.66 & \tabgain{+0.61} & 60.73 & 62.33 & \tabgain{+1.60} & 34.86 & 37.15 & \tabgain{+2.29} \\
        V2X-ViT & 72.24 & 72.31 & \tabgain{+0.07} & 62.89 & 64.76 & \tabgain{+1.87} & 36.02 & 38.01 & \tabgain{+1.99} \\
        V2VNet & 57.57 & 58.48 & \tabgain{+0.92} & 48.06 & 48.80 & \tabgain{+0.75} & 24.13 & 25.25 & \tabgain{+1.12} \\
        DiscoNet & 68.26 & 69.67 & \tabgain{+1.40} & 59.20 & 61.39 & \tabgain{+2.19} & 34.30 & 35.80 & \tabgain{+1.50} \\
        CoAlign & 61.59 & 61.39 & \tabdrop{-0.20} & 50.01 & 50.02 & \tabgain{+0.01} & 25.14 & 25.43 & \tabgain{+0.29} \\
        ERMVP & 67.43 & 67.25 & \tabdrop{-0.18} & 57.81 & 58.03 & \tabgain{+0.22} & 32.01 & 33.62 & \tabgain{+1.61} \\
        HEAL & 66.99 & 67.54 & \tabgain{+0.55} & 58.98 & 59.55 & \tabgain{+0.57} & 33.55 & 34.85 & \tabgain{+1.30} \\
        S-AdaFusion & 65.66 & 66.42 & \tabgain{+0.76} & 54.89 & 56.07 & \tabgain{+1.19} & 27.93 & 29.10 & \tabgain{+1.17} \\
        \bottomrule
    \end{tabular*}
    \end{adjustbox}
    \vspace{3pt}
    \textbf{(b) Heavy Vehicle}\par\vspace{1pt}
    \begin{adjustbox}{max width=\textwidth}
    \begin{tabular*}{\textwidth}{@{\extracolsep{\fill}}l rrr rrr rrr@{}}
        \toprule
        \multirow{2}{*}{\textbf{Fusion method}} & \multicolumn{3}{c}{\textbf{AP@0.3}} &
        \multicolumn{3}{c}{\textbf{AP@0.5}} & \multicolumn{3}{c}{\textbf{AP@0.7}} \\
        \cmidrule(lr){2-4}\cmidrule(lr){5-7}\cmidrule(lr){8-10}
        & Native & $+$Geo & $\Delta$ & Native & $+$Geo & $\Delta$ & Native & $+$Geo & $\Delta$ \\
        \midrule
        AttFuse & 21.09 & 21.24 & \tabgain{+0.15} & 11.95 & 10.85 & \tabdrop{-1.09} & 2.28 & 1.48 & \tabdrop{-0.80} \\
        F-Cooper & 26.29 & 27.57 & \tabgain{+1.28} & 19.16 & 21.18 & \tabgain{+2.02} & 6.04 & 7.38 & \tabgain{+1.34} \\
        CoBEVT & 23.13 & 21.63 & \tabdrop{-1.50} & 16.78 & 15.67 & \tabdrop{-1.11} & 1.88 & 1.79 & \tabdrop{-0.09} \\
        Where2comm & 26.14 & 27.77 & \tabgain{+1.63} & 20.02 & 22.46 & \tabgain{+2.44} & 6.72 & 6.88 & \tabgain{+0.16} \\
        V2X-ViT & 30.11 & 34.77 & \tabgain{+4.66} & 19.16 & 23.54 & \tabgain{+4.38} & 5.86 & 8.02 & \tabgain{+2.16} \\
        V2VNet & 13.89 & 14.49 & \tabgain{+0.60} & 9.44 & 10.18 & \tabgain{+0.74} & 1.51 & 1.73 & \tabgain{+0.22} \\
        DiscoNet & 29.97 & 27.63 & \tabdrop{-2.34} & 25.09 & 21.97 & \tabdrop{-3.12} & 7.95 & 6.95 & \tabdrop{-1.00} \\
        CoAlign & 25.92 & 25.88 & \tabdrop{-0.05} & 17.50 & 17.87 & \tabgain{+0.37} & 3.59 & 3.56 & \tabdrop{-0.03} \\
        ERMVP & 24.21 & 26.61 & \tabgain{+2.40} & 18.94 & 19.75 & \tabgain{+0.81} & 5.89 & 6.10 & \tabgain{+0.21} \\
        HEAL & 15.12 & 21.06 & \tabgain{+5.95} & 10.16 & 15.91 & \tabgain{+5.75} & 1.41 & 1.21 & \tabdrop{-0.20} \\
        S-AdaFusion & 26.44 & 26.04 & \tabdrop{-0.40} & 15.37 & 14.59 & \tabdrop{-0.78} & 0.96 & 1.14 & \tabgain{+0.19} \\
        \bottomrule
    \end{tabular*}
    \end{adjustbox}
    \vspace{3pt}
    \textbf{(c) 2Wheeler}\par\vspace{1pt}
    \begin{adjustbox}{max width=\textwidth}
    \begin{tabular*}{\textwidth}{@{\extracolsep{\fill}}l rrr rrr rrr@{}}
        \toprule
        \multirow{2}{*}{\textbf{Fusion method}} & \multicolumn{3}{c}{\textbf{AP@0.3}} &
        \multicolumn{3}{c}{\textbf{AP@0.5}} & \multicolumn{3}{c}{\textbf{AP@0.7}} \\
        \cmidrule(lr){2-4}\cmidrule(lr){5-7}\cmidrule(lr){8-10}
        & Native & $+$Geo & $\Delta$ & Native & $+$Geo & $\Delta$ & Native & $+$Geo & $\Delta$ \\
        \midrule
        AttFuse & 7.98 & 10.67 & \tabgain{+2.70} & 6.04 & 8.74 & \tabgain{+2.69} & 1.38 & 1.68 & \tabgain{+0.30} \\
        F-Cooper & 14.44 & 16.87 & \tabgain{+2.42} & 10.97 & 13.04 & \tabgain{+2.07} & 2.64 & 3.49 & \tabgain{+0.85} \\
        CoBEVT & 13.51 & 15.62 & \tabgain{+2.11} & 10.96 & 13.15 & \tabgain{+2.19} & 2.38 & 2.74 & \tabgain{+0.37} \\
        Where2comm & 10.76 & 11.99 & \tabgain{+1.24} & 9.07 & 9.78 & \tabgain{+0.72} & 2.28 & 2.32 & \tabgain{+0.04} \\
        V2X-ViT & 16.91 & 20.00 & \tabgain{+3.08} & 13.50 & 16.57 & \tabgain{+3.07} & 2.08 & 3.44 & \tabgain{+1.36} \\
        V2VNet & 9.76 & 10.96 & \tabgain{+1.20} & 5.87 & 7.57 & \tabgain{+1.70} & 0.80 & 1.12 & \tabgain{+0.32} \\
        DiscoNet & 11.74 & 12.47 & \tabgain{+0.73} & 10.31 & 10.65 & \tabgain{+0.34} & 2.74 & 3.11 & \tabgain{+0.37} \\
        CoAlign & 15.91 & 16.24 & \tabgain{+0.34} & 7.89 & 8.01 & \tabgain{+0.12} & 0.61 & 0.63 & \tabgain{+0.02} \\
        ERMVP & 12.98 & 12.05 & \tabdrop{-0.93} & 10.35 & 9.42 & \tabdrop{-0.93} & 1.73 & 2.53 & \tabgain{+0.79} \\
        HEAL & 11.88 & 10.12 & \tabdrop{-1.76} & 7.39 & 7.10 & \tabdrop{-0.29} & 1.36 & 1.34 & \tabdrop{-0.02} \\
        S-AdaFusion & 14.68 & 15.97 & \tabgain{+1.29} & 8.83 & 8.60 & \tabdrop{-0.23} & 1.22 & 1.33 & \tabgain{+0.11} \\
        \bottomrule
    \end{tabular*}
    \end{adjustbox}
    \vspace{3pt}
    \textbf{(d) Pedestrian}\par\vspace{1pt}
    \begin{adjustbox}{max width=\textwidth}
    \begin{tabular*}{\textwidth}{@{\extracolsep{\fill}}l rrr rrr rrr@{}}
        \toprule
        \multirow{2}{*}{\textbf{Fusion method}} & \multicolumn{3}{c}{\textbf{AP@0.3}} &
        \multicolumn{3}{c}{\textbf{AP@0.5}} & \multicolumn{3}{c}{\textbf{AP@0.7}} \\
        \cmidrule(lr){2-4}\cmidrule(lr){5-7}\cmidrule(lr){8-10}
        & Native & $+$Geo & $\Delta$ & Native & $+$Geo & $\Delta$ & Native & $+$Geo & $\Delta$ \\
        \midrule
        AttFuse & 9.72 & 9.73 & \tabgain{+0.01} & 6.07 & 6.36 & \tabgain{+0.29} & 0.36 & 0.34 & \tabdrop{-0.02} \\
        F-Cooper & 15.42 & 16.13 & \tabgain{+0.71} & 10.08 & 11.37 & \tabgain{+1.29} & 0.59 & 0.92 & \tabgain{+0.33} \\
        CoBEVT & 11.80 & 11.88 & \tabgain{+0.07} & 6.76 & 6.69 & \tabdrop{-0.07} & 0.40 & 0.46 & \tabgain{+0.05} \\
        Where2comm & 14.90 & 14.17 & \tabdrop{-0.73} & 9.76 & 10.06 & \tabgain{+0.30} & 0.66 & 0.76 & \tabgain{+0.10} \\
        V2X-ViT & 16.17 & 18.33 & \tabgain{+2.17} & 9.96 & 11.47 & \tabgain{+1.51} & 0.70 & 0.88 & \tabgain{+0.17} \\
        V2VNet & 10.13 & 9.99 & \tabdrop{-0.14} & 5.10 & 5.27 & \tabgain{+0.17} & 0.15 & 0.20 & \tabgain{+0.05} \\
        DiscoNet & 12.58 & 13.61 & \tabgain{+1.03} & 7.46 & 9.00 & \tabgain{+1.54} & 0.47 & 0.60 & \tabgain{+0.12} \\
        CoAlign & 9.45 & 9.56 & \tabgain{+0.12} & 4.43 & 4.51 & \tabgain{+0.08} & 0.16 & 0.17 & \tabgain{+0.01} \\
        ERMVP & 12.14 & 13.53 & \tabgain{+1.39} & 7.94 & 9.41 & \tabgain{+1.47} & 0.53 & 0.64 & \tabgain{+0.11} \\
        HEAL & 12.39 & 13.74 & \tabgain{+1.36} & 6.34 & 7.54 & \tabgain{+1.20} & 0.18 & 0.25 & \tabgain{+0.08} \\
        S-AdaFusion & 11.46 & 11.53 & \tabgain{+0.07} & 6.61 & 6.67 & \tabgain{+0.06} & 0.35 & 0.39 & \tabgain{+0.04} \\
        \bottomrule
    \end{tabular*}
    \end{adjustbox}
\end{table*}
\begin{table*}[p]
    \centering
    \caption{Distance-stratified analysis on the UrbanIng-V2X crossing1/2 validation split. Each
    panel uses the evaluator's native range partition and reports class-macro AP for Native,
    $+$Geo, and their signed percentage-point difference. \textcolor{ForestGreen}{Green} and
    \textcolor{BrickRed}{red} $\Delta$ cells mark gains and drops.}
    \label{tab:distance-range-analysis}
    \scriptsize
    \setlength{\tabcolsep}{2.8pt}
    \renewcommand{\arraystretch}{0.98}
    \textbf{(a) Near range: 0--30 m}\par\vspace{1pt}
    \begin{adjustbox}{max width=\textwidth}
    \begin{tabular*}{\textwidth}{@{\extracolsep{\fill}}l rrr rrr rrr@{}}
        \toprule
        \multirow{2}{*}{\textbf{Fusion method}} & \multicolumn{3}{c}{\textbf{AP@0.3}} &
        \multicolumn{3}{c}{\textbf{AP@0.5}} & \multicolumn{3}{c}{\textbf{AP@0.7}} \\
        \cmidrule(lr){2-4}\cmidrule(lr){5-7}\cmidrule(lr){8-10}
        & Native & $+$Geo & $\Delta$ & Native & $+$Geo & $\Delta$ & Native & $+$Geo & $\Delta$ \\
        \midrule
        AttFuse & 37.46 & 37.89 & \tabgain{+0.44} & 28.86 & 29.72 & \tabgain{+0.86} & 13.61 & 13.39 & \tabdrop{-0.23} \\
        F-Cooper & 47.53 & 47.18 & \tabdrop{-0.35} & 41.26 & 38.48 & \tabdrop{-2.79} & 20.24 & 21.59 & \tabgain{+1.36} \\
        CoBEVT & 39.75 & 38.92 & \tabdrop{-0.83} & 35.56 & 34.76 & \tabdrop{-0.80} & 16.36 & 15.94 & \tabdrop{-0.42} \\
        Where2comm & 45.58 & 49.42 & \tabgain{+3.85} & 39.41 & 45.09 & \tabgain{+5.68} & 21.05 & 24.27 & \tabgain{+3.22} \\
        V2X-ViT & 49.44 & 52.83 & \tabgain{+3.38} & 40.79 & 44.80 & \tabgain{+4.00} & 20.32 & 21.67 & \tabgain{+1.35} \\
        V2VNet & 33.02 & 33.74 & \tabgain{+0.72} & 27.26 & 27.98 & \tabgain{+0.72} & 13.36 & 14.11 & \tabgain{+0.75} \\
        DiscoNet & 46.62 & 46.36 & \tabdrop{-0.26} & 42.53 & 40.81 & \tabdrop{-1.71} & 21.48 & 21.11 & \tabdrop{-0.37} \\
        CoAlign & 43.20 & 43.87 & \tabgain{+0.67} & 33.94 & 35.21 & \tabgain{+1.27} & 14.60 & 14.85 & \tabgain{+0.25} \\
        ERMVP & 43.32 & 44.21 & \tabgain{+0.89} & 37.25 & 38.77 & \tabgain{+1.53} & 19.61 & 21.33 & \tabgain{+1.72} \\
        HEAL & 38.55 & 41.81 & \tabgain{+3.27} & 31.39 & 36.37 & \tabgain{+4.97} & 16.41 & 17.03 & \tabgain{+0.61} \\
        S-AdaFusion & 44.46 & 44.68 & \tabgain{+0.23} & 34.33 & 34.07 & \tabdrop{-0.26} & 14.11 & 14.73 & \tabgain{+0.61} \\
        \bottomrule
    \end{tabular*}
    \end{adjustbox}
    \vspace{4pt}
    \textbf{(b) Middle range: 30--50 m}\par\vspace{1pt}
    \begin{adjustbox}{max width=\textwidth}
    \begin{tabular*}{\textwidth}{@{\extracolsep{\fill}}l rrr rrr rrr@{}}
        \toprule
        \multirow{2}{*}{\textbf{Fusion method}} & \multicolumn{3}{c}{\textbf{AP@0.3}} &
        \multicolumn{3}{c}{\textbf{AP@0.5}} & \multicolumn{3}{c}{\textbf{AP@0.7}} \\
        \cmidrule(lr){2-4}\cmidrule(lr){5-7}\cmidrule(lr){8-10}
        & Native & $+$Geo & $\Delta$ & Native & $+$Geo & $\Delta$ & Native & $+$Geo & $\Delta$ \\
        \midrule
        AttFuse & 24.83 & 25.72 & \tabgain{+0.89} & 18.26 & 19.64 & \tabgain{+1.38} & 7.29 & 7.32 & \tabgain{+0.03} \\
        F-Cooper & 34.07 & 35.98 & \tabgain{+1.91} & 27.50 & 29.39 & \tabgain{+1.88} & 12.21 & 13.95 & \tabgain{+1.74} \\
        CoBEVT & 29.49 & 31.03 & \tabgain{+1.54} & 23.64 & 25.66 & \tabgain{+2.02} & 9.29 & 9.77 & \tabgain{+0.48} \\
        Where2comm & 31.09 & 31.84 & \tabgain{+0.75} & 26.52 & 27.66 & \tabgain{+1.14} & 12.47 & 12.76 & \tabgain{+0.30} \\
        V2X-ViT & 37.30 & 39.99 & \tabgain{+2.69} & 29.67 & 32.65 & \tabgain{+2.98} & 12.11 & 13.79 & \tabgain{+1.67} \\
        V2VNet & 23.34 & 24.35 & \tabgain{+1.01} & 16.96 & 18.14 & \tabgain{+1.18} & 6.65 & 7.01 & \tabgain{+0.36} \\
        DiscoNet & 32.16 & 31.62 & \tabdrop{-0.54} & 27.12 & 27.08 & \tabdrop{-0.04} & 13.43 & 13.22 & \tabdrop{-0.21} \\
        CoAlign & 32.83 & 33.01 & \tabgain{+0.18} & 22.05 & 22.31 & \tabgain{+0.26} & 7.62 & 7.69 & \tabgain{+0.07} \\
        ERMVP & 33.17 & 32.32 & \tabdrop{-0.85} & 27.66 & 26.62 & \tabdrop{-1.03} & 11.60 & 12.12 & \tabgain{+0.52} \\
        HEAL & 28.01 & 29.10 & \tabgain{+1.09} & 21.02 & 22.31 & \tabgain{+1.30} & 8.96 & 9.23 & \tabgain{+0.27} \\
        S-AdaFusion & 29.06 & 29.86 & \tabgain{+0.80} & 20.86 & 20.68 & \tabdrop{-0.18} & 7.72 & 8.03 & \tabgain{+0.31} \\
        \bottomrule
    \end{tabular*}
    \end{adjustbox}
    \vspace{4pt}
    \textbf{(c) Far range: 50--100 m}\par\vspace{1pt}
    \begin{adjustbox}{max width=\textwidth}
    \begin{tabular*}{\textwidth}{@{\extracolsep{\fill}}l rrr rrr rrr@{}}
        \toprule
        \multirow{2}{*}{\textbf{Fusion method}} & \multicolumn{3}{c}{\textbf{AP@0.3}} &
        \multicolumn{3}{c}{\textbf{AP@0.5}} & \multicolumn{3}{c}{\textbf{AP@0.7}} \\
        \cmidrule(lr){2-4}\cmidrule(lr){5-7}\cmidrule(lr){8-10}
        & Native & $+$Geo & $\Delta$ & Native & $+$Geo & $\Delta$ & Native & $+$Geo & $\Delta$ \\
        \midrule
        AttFuse & 18.74 & 19.34 & \tabgain{+0.60} & 13.22 & 13.47 & \tabgain{+0.25} & 4.55 & 4.38 & \tabdrop{-0.16} \\
        F-Cooper & 24.22 & 25.79 & \tabgain{+1.57} & 17.58 & 20.55 & \tabgain{+2.97} & 6.66 & 7.81 & \tabgain{+1.15} \\
        CoBEVT & 23.47 & 23.38 & \tabdrop{-0.08} & 16.62 & 16.75 & \tabgain{+0.13} & 5.11 & 5.38 & \tabgain{+0.27} \\
        Where2comm & 23.27 & 23.53 & \tabgain{+0.26} & 17.62 & 18.15 & \tabgain{+0.52} & 6.13 & 6.45 & \tabgain{+0.32} \\
        V2X-ViT & 25.99 & 28.03 & \tabgain{+2.04} & 18.26 & 20.49 & \tabgain{+2.23} & 6.49 & 7.63 & \tabgain{+1.14} \\
        V2VNet & 16.03 & 16.49 & \tabgain{+0.46} & 10.73 & 11.33 & \tabgain{+0.60} & 2.81 & 3.16 & \tabgain{+0.34} \\
        DiscoNet & 23.67 & 23.85 & \tabgain{+0.18} & 18.20 & 18.41 & \tabgain{+0.21} & 6.57 & 6.73 & \tabgain{+0.16} \\
        CoAlign & 19.25 & 19.17 & \tabdrop{-0.09} & 12.60 & 12.62 & \tabgain{+0.02} & 3.98 & 3.96 & \tabdrop{-0.02} \\
        ERMVP & 20.94 & 21.78 & \tabgain{+0.84} & 15.62 & 15.97 & \tabgain{+0.36} & 5.31 & 5.63 & \tabgain{+0.33} \\
        HEAL & 20.03 & 21.90 & \tabgain{+1.87} & 14.84 & 16.45 & \tabgain{+1.61} & 5.28 & 5.43 & \tabgain{+0.15} \\
        S-AdaFusion & 24.30 & 24.99 & \tabgain{+0.69} & 16.04 & 16.42 & \tabgain{+0.37} & 4.29 & 4.65 & \tabgain{+0.36} \\
        \bottomrule
    \end{tabular*}
    \end{adjustbox}
\end{table*}

\subsection{Complete Generalization Results}
\label{sec:complete-generalization}
Tables~\ref{tab:crossing3-generalization} and~\ref{tab:v2v4real-transfer} extend
Table~\ref{tab:generalization} with AP@0.3 and the Ego-only, Early Fusion, and Late Fusion
detectors on crossing3 and V2V4Real.

\begin{table*}[t]
    \centering
    \caption{Generalization to the held-out UrbanIng-V2X intersection (crossing3). Final checkpoints are
    evaluated without retraining or checkpoint selection. Reference detectors report
    Native AP only; for each fusion method, $\Delta$ is $\text{GeoScaffold}-\text{Native}$.
    AP is in percent and $\Delta$ in percentage points; \textcolor{ForestGreen}{green} and
    \textcolor{BrickRed}{red} cells mark gains and drops.}
    \label{tab:crossing3-generalization}
    \small
    \renewcommand{\arraystretch}{1.08}
    \begin{tabular*}{\textwidth}{@{\extracolsep{\fill}}l rrr rrr rrr@{}}
        \toprule
        \multirow{2}{*}{\textbf{Fusion method}} & \multicolumn{3}{c}{\textbf{AP@0.3}} &
        \multicolumn{3}{c}{\textbf{AP@0.5}} & \multicolumn{3}{c}{\textbf{AP@0.7}} \\
        \cmidrule(lr){2-4}\cmidrule(lr){5-7}\cmidrule(lr){8-10}
        & Native & $+$Geo & $\Delta$ & Native & $+$Geo & $\Delta$ & Native & $+$Geo & $\Delta$ \\
        \midrule
        \multicolumn{10}{@{}l}{\textit{Reference detectors}} \\
        Ego-only & 15.51 & \tabna & \tabna & 12.18 & \tabna & \tabna & 3.75 & \tabna & \tabna \\
        Early Fusion & 17.59 & \tabna & \tabna & 14.61 & \tabna & \tabna & 5.87 & \tabna & \tabna \\
        Late Fusion & 12.01 & \tabna & \tabna & 9.18 & \tabna & \tabna & 3.37 & \tabna & \tabna \\
        \midrule
        \multicolumn{10}{@{}l}{\textit{Fusion methods with Native/$+$Geo pairs}} \\
        AttFuse & 23.11 & 23.80 & \tabgain{+0.69} & 19.75 & 20.05 & \tabgain{+0.30} & 7.43 & 7.90 & \tabgain{+0.47} \\
        F-Cooper & 26.10 & 32.61 & \tabgain{+6.51} & 22.70 & 29.39 & \tabgain{+6.69} & 9.98 & 13.83 & \tabgain{+3.85} \\
        CoBEVT & 23.78 & 24.71 & \tabgain{+0.92} & 20.70 & 21.52 & \tabgain{+0.83} & 9.20 & 9.49 & \tabgain{+0.29} \\
        Where2comm & 26.27 & 27.39 & \tabgain{+1.12} & 22.48 & 23.41 & \tabgain{+0.93} & 10.65 & 11.08 & \tabgain{+0.43} \\
        V2X-ViT & 25.70 & 27.68 & \tabgain{+1.99} & 21.99 & 22.91 & \tabgain{+0.92} & 9.67 & 10.01 & \tabgain{+0.34} \\
        V2VNet & 19.69 & 20.18 & \tabgain{+0.49} & 15.37 & 15.76 & \tabgain{+0.39} & 6.01 & 6.12 & \tabgain{+0.11} \\
        DiscoNet & 27.70 & 28.22 & \tabgain{+0.52} & 22.90 & 24.12 & \tabgain{+1.23} & 11.14 & 13.20 & \tabgain{+2.07} \\
        CoAlign & 17.78 & 17.63 & \tabdrop{-0.14} & 13.97 & 13.74 & \tabdrop{-0.23} & 4.91 & 4.87 & \tabdrop{-0.03} \\
        ERMVP & 25.55 & 27.02 & \tabgain{+1.47} & 22.67 & 24.45 & \tabgain{+1.78} & 11.08 & 10.37 & \tabdrop{-0.71} \\
        HEAL & 15.96 & 17.72 & \tabgain{+1.75} & 13.54 & 13.90 & \tabgain{+0.36} & 4.48 & 4.90 & \tabgain{+0.43} \\
        S-AdaFusion & 23.21 & 22.85 & \tabdrop{-0.36} & 19.86 & 19.58 & \tabdrop{-0.28} & 7.05 & 7.04 & \tabdrop{-0.01} \\
        \bottomrule
    \end{tabular*}
\end{table*}
\begin{table*}[t]
    \centering
    \caption{Results on the V2V4Real test split. Every Native/$+$Geo pair is
    trained from scratch on V2V4Real with matched schedules, seeds, augmentation, and data
    order (App.~\ref{sec:experimental-protocols}) and evaluated with the dataset's single-class
    vehicle BEV evaluator. AP is in percent and
    $\Delta=\text{GeoScaffold}-\text{Native}$ in percentage points;
    \textcolor{ForestGreen}{green} and \textcolor{BrickRed}{red} cells mark gains and drops.}
    \label{tab:v2v4real-transfer}
    \small
    \renewcommand{\arraystretch}{1.08}
    \begin{tabular*}{\textwidth}{@{\extracolsep{\fill}}l rrr rrr rrr@{}}
        \toprule
        \multirow{2}{*}{\textbf{Fusion method}} &
        \multicolumn{3}{c}{\textbf{AP@0.3}} &
        \multicolumn{3}{c}{\textbf{AP@0.5}} &
        \multicolumn{3}{c}{\textbf{AP@0.7}} \\
        \cmidrule(lr){2-4}\cmidrule(lr){5-7}\cmidrule(lr){8-10}
        & Native & $+$Geo & $\Delta$ & Native & $+$Geo & $\Delta$ &
        Native & $+$Geo & $\Delta$ \\
        \midrule
        \multicolumn{10}{@{}l}{\textit{Reference detectors}} \\
        Ego-only & 52.29 & \tabna & \tabna & 47.29 & \tabna & \tabna & 31.06 & \tabna & \tabna \\
        Early Fusion & 68.54 & \tabna & \tabna & 60.33 & \tabna & \tabna & 31.14 & \tabna & \tabna \\
        Late Fusion & 73.61 & \tabna & \tabna & 65.41 & \tabna & \tabna & 38.08 & \tabna & \tabna \\
        \midrule
        \multicolumn{10}{@{}l}{\textit{Fusion methods with Native/$+$Geo pairs}} \\
        AttFuse & 70.11 & 73.98 & \tabgain{+3.88} & 63.71 & 67.28 & \tabgain{+3.57} & 37.41 & 38.38 & \tabgain{+0.97} \\
        F-Cooper & 71.05 & 72.81 & \tabgain{+1.76} & 63.44 & 66.45 & \tabgain{+3.01} & 36.29 & 38.41 & \tabgain{+2.13} \\
        CoBEVT & 72.70 & 73.27 & \tabgain{+0.56} & 67.12 & 69.08 & \tabgain{+1.96} & 39.39 & 43.76 & \tabgain{+4.36} \\
        Where2comm & 68.42 & 68.44 & \tabgain{+0.02} & 59.47 & 61.29 & \tabgain{+1.83} & 31.68 & 34.50 & \tabgain{+2.83} \\
        V2X-ViT & 73.48 & 73.10 & \tabdrop{-0.38} & 67.11 & 67.81 & \tabgain{+0.71} & 36.35 & 39.74 & \tabgain{+3.39} \\
        V2VNet$^{\dagger}$ & \textendash & \textendash & \textendash & \textendash & \textendash & \textendash & \textendash & \textendash & \textendash \\
        DiscoNet & 64.02 & 66.59 & \tabgain{+2.57} & 54.58 & 58.55 & \tabgain{+3.96} & 27.15 & 28.81 & \tabgain{+1.66} \\
        CoAlign & 65.35 & 65.60 & \tabgain{+0.25} & 50.47 & 53.22 & \tabgain{+2.74} & 26.68 & 27.94 & \tabgain{+1.27} \\
        ERMVP & 75.02 & 74.75 & \tabdrop{-0.26} & 67.34 & 67.81 & \tabgain{+0.48} & 41.59 & 42.41 & \tabgain{+0.82} \\
        HEAL & 76.26 & 75.95 & \tabdrop{-0.31} & 69.89 & 70.54 & \tabgain{+0.65} & 37.19 & 40.13 & \tabgain{+2.94} \\
        S-AdaFusion & 69.68 & 69.95 & \tabgain{+0.27} & 61.95 & 62.09 & \tabgain{+0.14} & 31.90 & 33.58 & \tabgain{+1.67} \\
        \bottomrule
    \end{tabular*}
    \vspace{3pt}
    \begin{minipage}{\textwidth}\footnotesize
        $^{\dagger}$V2VNet (Native, $+$Geo) produced no detections, so no AP is reported.
    \end{minipage}
\end{table*}

\subsection{Pose Error and Latency}
\label{sec:pose-latency}
Fig.~\ref{fig:robustness} evaluates five fusion methods under test-time pose noise and communication
latency on the crossing1/2 validation split. Pose noise adds independent zero-mean Gaussian perturbations to the $x$, $y$, and $z$
position and to the yaw of each collaborator's pose, with standard deviations of 0.2, 0.4, and
0.6 (meters and degrees); the receiver's pose is unchanged. Under latency, a delayed collaborator
contributes its newest LiDAR sweep and pose that are at least the requested delay (100 to
500\,ms) older than the receiver frame. At the start of a sequence, where no such sweep exists,
its earliest sweep is used, so the realized delay averages 492.5\,ms at the requested 500\,ms.
Both arms of a pair receive the same perturbation for each frame and collaborator.

\IfFileExists{figures/robustness/pose_latency_robustness.pdf}{%
\begin{figure*}[tp]
    \centering
    \includegraphics[width=0.93\textwidth]{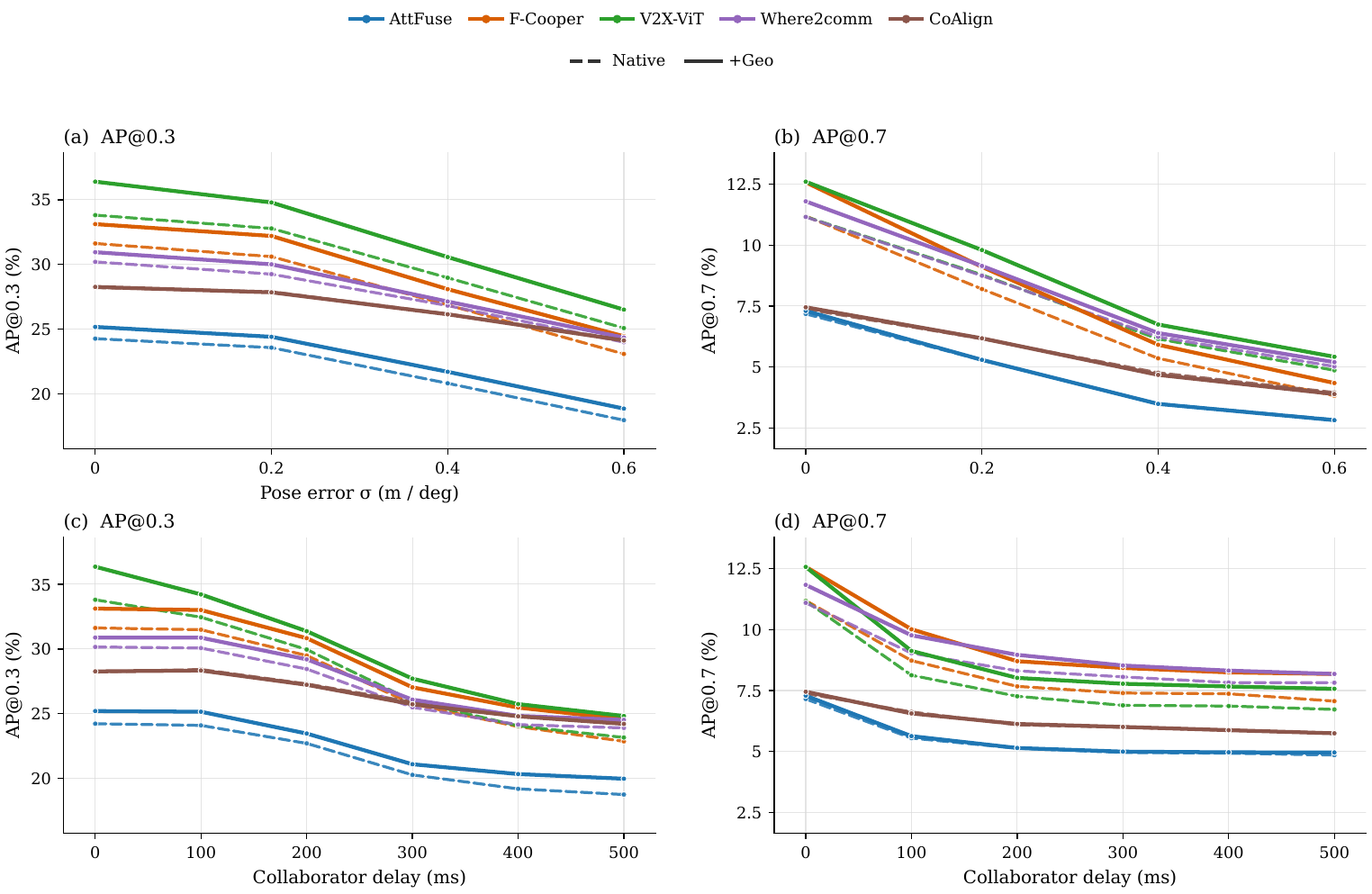}
    \caption{\textbf{Robustness under test-time pose error (top) and communication latency (bottom).}
    Columns report AP@0.3 and AP@0.7. Each color denotes a fusion method; dashed and solid
    curves denote Native and $+$Geo, respectively. Every paired comparison uses the same
    frame--agent perturbations or causal historical observations, and AP is reported in percent.
    Pairs use the checkpoints of Table~\ref{tab:main-results}.}
    \label{fig:robustness}
\end{figure*}
}{}

\subsection{Design Controls}
\label{sec:design-controls}
Table~\ref{tab:design-crossing3} evaluates the F-Cooper arms of
Table~\ref{tab:geometry-ablation}(a) and the V2X-ViT controls of
Table~\ref{tab:geometry-ablation}(b)--(c) on crossing3 with the same final checkpoints.
Table~\ref{tab:attfuse-ablation} repeats the controls of Table~\ref{tab:geometry-ablation}(b)--(d)
on AttFuse under its matched 20-epoch protocol. In both tables, every auxiliary and setting
improves over Native at all three thresholds.

\begin{table*}[p]
    \centering
    \caption{Design controls of Table~\ref{tab:geometry-ablation} on the held-out
    crossing3 intersection, evaluated with the same final checkpoints: the F-Cooper arms of
    panel (a) and the V2X-ViT controls of panels (b)--(c). Crossing3 is not used to select
    settings. \colorbox{tableAccent}{Blue} shading marks the default; AP is reported in
    percent.}
    \label{tab:design-crossing3}
    \footnotesize
    \setlength{\tabcolsep}{4pt}
    \renewcommand{\arraystretch}{1.06}
    \begin{tabular}{@{}l rrr@{}}
        \toprule
        \textbf{Training setting} & AP@0.3 & AP@0.5 & AP@0.7 \\
        \midrule
        \multicolumn{4}{@{}l}{\textit{(a) Representation and objective, F-Cooper}} \\
        None (Native) & 26.10 & 22.70 & 9.98 \\
        7-D box code, SmoothL1 & 30.78 & 25.76 & 11.83 \\
        Foreground mask, CE & 29.73 & 25.09 & 10.86 \\
        Foreground mask, Dice & 29.03 & 23.80 & 11.26 \\
        \rowcolor{tableAccent}
        Foreground mask, CE $+$ Dice & 32.61 & 29.39 & 13.83 \\
        \midrule
        \multicolumn{4}{@{}l}{\textit{(b)--(c) Supervised feature and sensitivity, V2X-ViT}} \\
        None (Native) & 25.70 & 21.99 & 9.67 \\
        Ego feature & 26.10 & 22.84 & 10.62 \\
        \rowcolor{tableAccent}
        Fused feature, $\lambda_{\mathrm{Geo}}{=}1$, $K{=}28$ & 27.68 & 22.91 & 10.01 \\
        Fused feature, $\lambda_{\mathrm{Geo}}{=}0.5$ & 29.40 & 24.14 & 10.59 \\
        Fused feature, $\lambda_{\mathrm{Geo}}{=}2$ & 29.49 & 24.80 & 10.69 \\
        Fused feature, $K{=}14$ & 28.96 & 25.46 & 10.49 \\
        Fused feature, $K{=}56$ & 26.67 & 22.73 & 9.69 \\
        \bottomrule
    \end{tabular}
\end{table*}

\begin{table*}[p]
    \centering
    \caption{GeoScaffold design controls on AttFuse (UrbanIng-V2X validation split, matched
    20-epoch AttFuse protocol). Panels mirror Table~\ref{tab:geometry-ablation}(b)--(d):
    \textbf{(a)} the feature supervised by the auxiliary, \textbf{(b)} one-at-a-time
    variation of $\lambda_{\mathrm{Geo}}$ and the mask resolution $K$ with CE and Dice at $1{:}1$,
    and \textbf{(c)} held-out soft IoU of the same frozen linear probe, averaged within each
    collaborative frame and then over frames. \colorbox{tableAccent}{Blue} shading marks the
    defaults; AP and soft IoU are reported in percent.}
    \label{tab:attfuse-ablation}
    \footnotesize
    \setlength{\tabcolsep}{3.2pt}
    \renewcommand{\arraystretch}{1.06}
    \begin{minipage}[t]{0.305\textwidth}
        \centering
        \textbf{(a) Supervised feature}\par\vspace{2pt}
        \begin{tabular}{@{}l rrr@{}}
            \toprule
            \textbf{Feature} & AP@0.3 & AP@0.5 & AP@0.7 \\
            \midrule
            None (Native) & 24.25 & 18.18 & 7.16 \\
            Ego feature & 24.47 & 18.50 & \textbf{7.41} \\
            \rowcolor{tableAccent}
            Fused feature & \textbf{25.19} & \textbf{18.89} & 7.30 \\
            \bottomrule
        \end{tabular}
    \end{minipage}\hfill
    \begin{minipage}[t]{0.385\textwidth}
        \centering
        \textbf{(b) Hyperparameter sensitivity}\par\vspace{2pt}
        \begin{tabular}{@{}l c rrr@{}}
            \toprule
            \textbf{Factor} & \textbf{Value} & AP@0.3 & AP@0.5 & AP@0.7 \\
            \midrule
            Loss weight $\lambda_{\mathrm{Geo}}$ & 0.5 & 25.28 & 19.10 & 7.57 \\
            \rowcolor{tableAccent}
            & 1 & 25.19 & 18.89 & 7.30 \\
            & 2 & 25.37 & 19.58 & 7.55 \\
            \midrule
            Mask resolution $K$ & 14 & 25.31 & 19.10 & 7.64 \\
            \rowcolor{tableAccent}
            & 28 & 25.19 & 18.89 & 7.30 \\
            & 56 & 25.11 & 19.27 & 7.55 \\
            \bottomrule
        \end{tabular}
    \end{minipage}\hfill
    \begin{minipage}[t]{0.27\textwidth}
        \centering
        \textbf{(c) Mask readability}\par\vspace{2pt}
        \begin{tabular}{@{}l r@{}}
            \toprule
            \textbf{Probe input} & \textbf{Soft IoU $\uparrow$} \\
            \midrule
            Class $+$ template prior & 59.86 \\
            Native fused ROI & 60.73 \\
            \rowcolor{tableAccent}
            $+$Geo fused ROI & \textbf{66.17} \\
            $+$Geo ROI, permuted & 53.66 \\
            \bottomrule
        \end{tabular}
    \end{minipage}
\end{table*}

\subsection{Mask Readability}
\label{sec:mask-readability}
\begin{figure*}[p]
    \centering
    \includegraphics[width=0.82\textwidth]{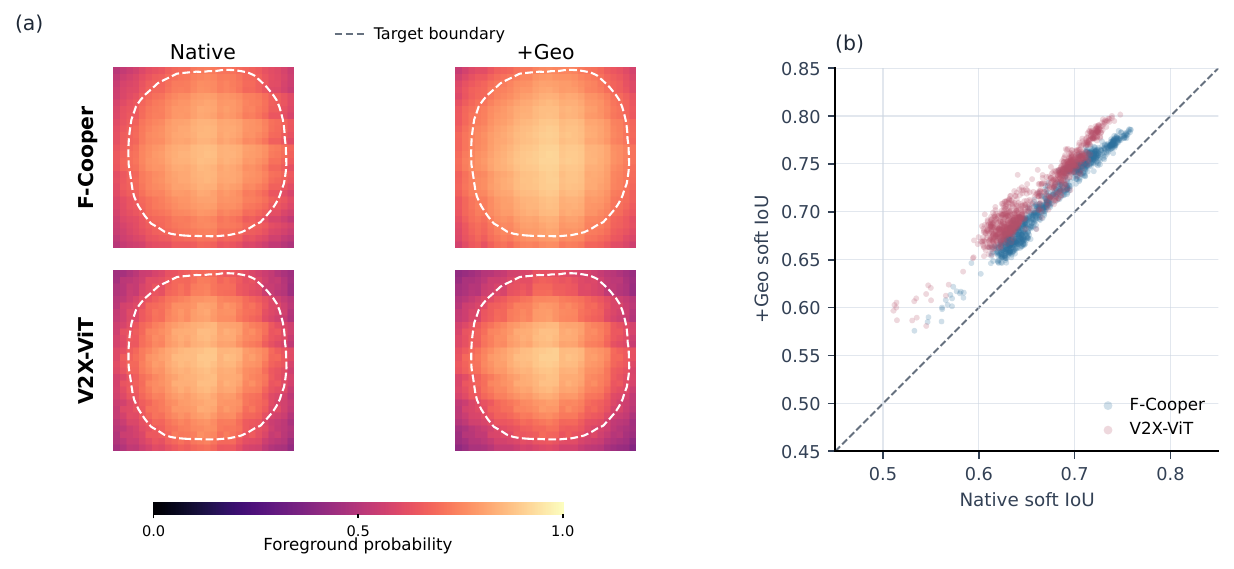}
    \caption{\textbf{Anchor-canonical foreground-mask probe.}
    Native and $+$Geo fused ROIs are decoded by the same linear probe.
    \textbf{(a)} Class-balanced mean probe probability maps over collaborative validation samples.
    Every map uses the same coordinates and probability scale; the dashed curve is the
    $0.5$ isocontour of the class-balanced mean foreground mask. \textbf{(b)} Directly paired frame-level
    soft IoU observations, with each point formed by averaging positive anchors from the same
    collaborative frame and the dashed line denoting equality. Controlled quantitative comparisons are reported in
    Table~\ref{tab:geometry-ablation}.}
    \label{fig:feature-distribution}
\end{figure*}

Fig.~\ref{fig:feature-distribution} visualizes the probe of
Table~\ref{tab:geometry-ablation}(d). For each pair, a single linear transposed convolution (kernel size and stride 4, mapping the
$7\times7$ ROI to $28\times28$ logits) is trained for one epoch on the Native and $+$Geo ROIs of
all training frames while both detectors stay frozen. Training uses AdamW with learning rate
$10^{-3}$ and weight decay $10^{-4}$, no data augmentation, and the average of the two arms'
CE $+$ Dice losses. The probe is then evaluated on the collaborative validation frames. For a positive anchor
$a$ with target $\tau_{a,pq}$ (Eq.~\eqref{eq:mask-loss}) and probe foreground probability
$\tilde{\mu}_{a,pq}$, the soft IoU is
\[
    \frac{\sum_{p,q}\tilde{\mu}_{a,pq}\,\tau_{a,pq}}
         {\sum_{p,q}\bigl(\tilde{\mu}_{a,pq}+\tau_{a,pq}-\tilde{\mu}_{a,pq}\,\tau_{a,pq}\bigr)}\in[0,1],
\]
which is averaged over the positive anchors of each frame and then over frames.

\clearpage
\onecolumn
\section{Qualitative Results}
\label{sec:qualitative}
Figs.~\ref{fig:qualitative-val-1}--\ref{fig:qualitative-test-2} compare the Native and $+$Geo
outputs of ten fusion methods on crossing1/2 and on the held-out crossing3 intersection.

\newcommand{\qualitativepage}[5][p]{%
\IfFileExists{#2}{%
\begin{figure*}[#1]
    \centering
    \includegraphics[width=0.95\textwidth]{#2}
    \caption{\textbf{#3.}
    Each row shows two scenes for one fusion method, each with its Native and $+$Geo outputs.
    Both outputs are drawn on the same raw LiDAR from that method's input agents, with the same
    ground-truth boxes, crop, and inset scale.
    Solid outlines are predictions;
    orange dashed outlines are ground truth. Thicker outlines identify a shared target
    and its same-class, one-to-one matched prediction. Neighboring predictions and targets
    remain visible with lighter outlines; no additional prediction threshold is applied.
    The scenes are illustrative and were selected to show visible differences.
    #4}
    \label{#5}
\end{figure*}
}{}%
}

\qualitativepage[!ht]{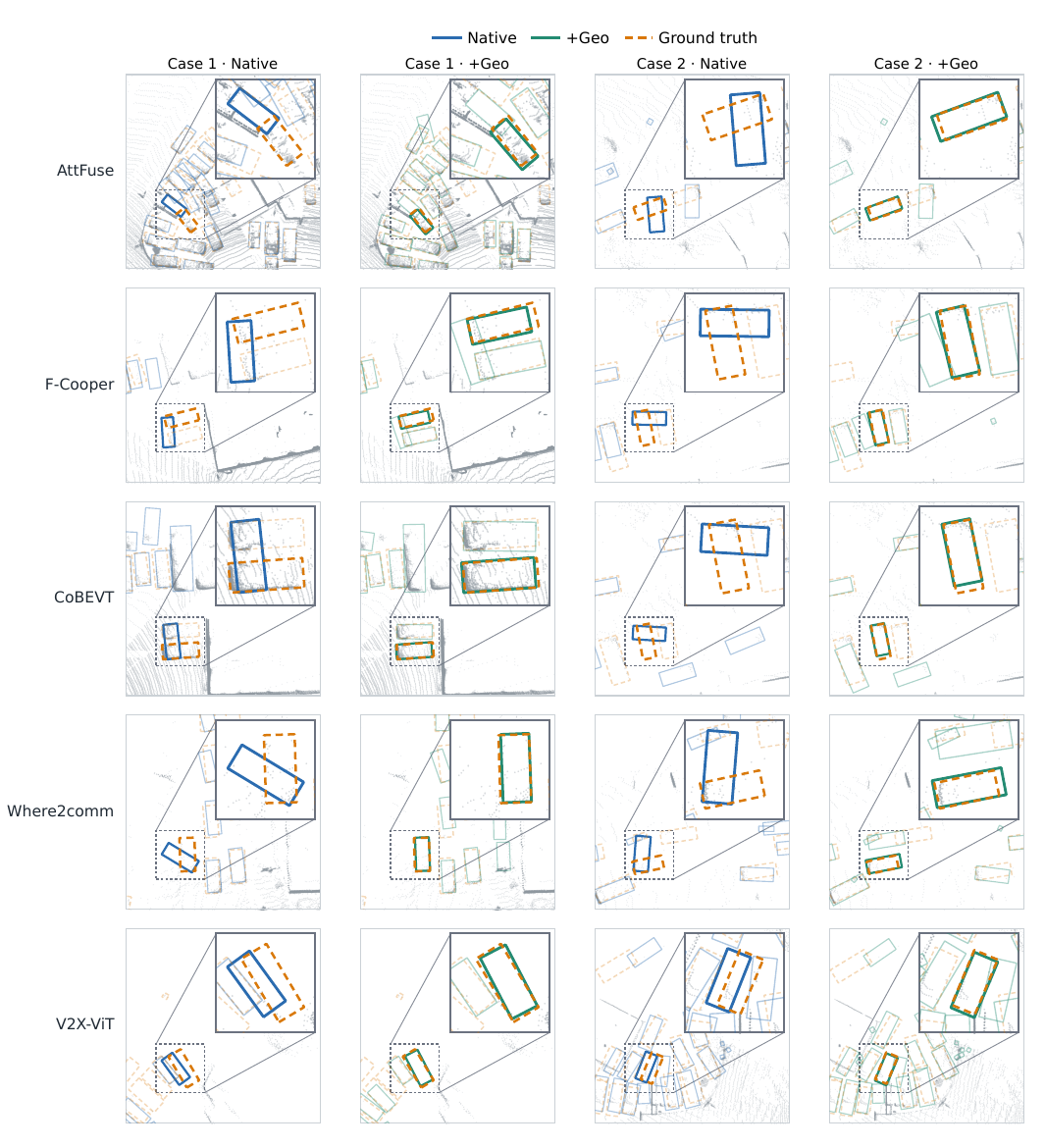}
    {Qualitative comparisons on crossing1/2}
    {The remaining methods are shown in Fig.~\ref{fig:qualitative-val-2}.}
    {fig:qualitative-val-1}
\qualitativepage{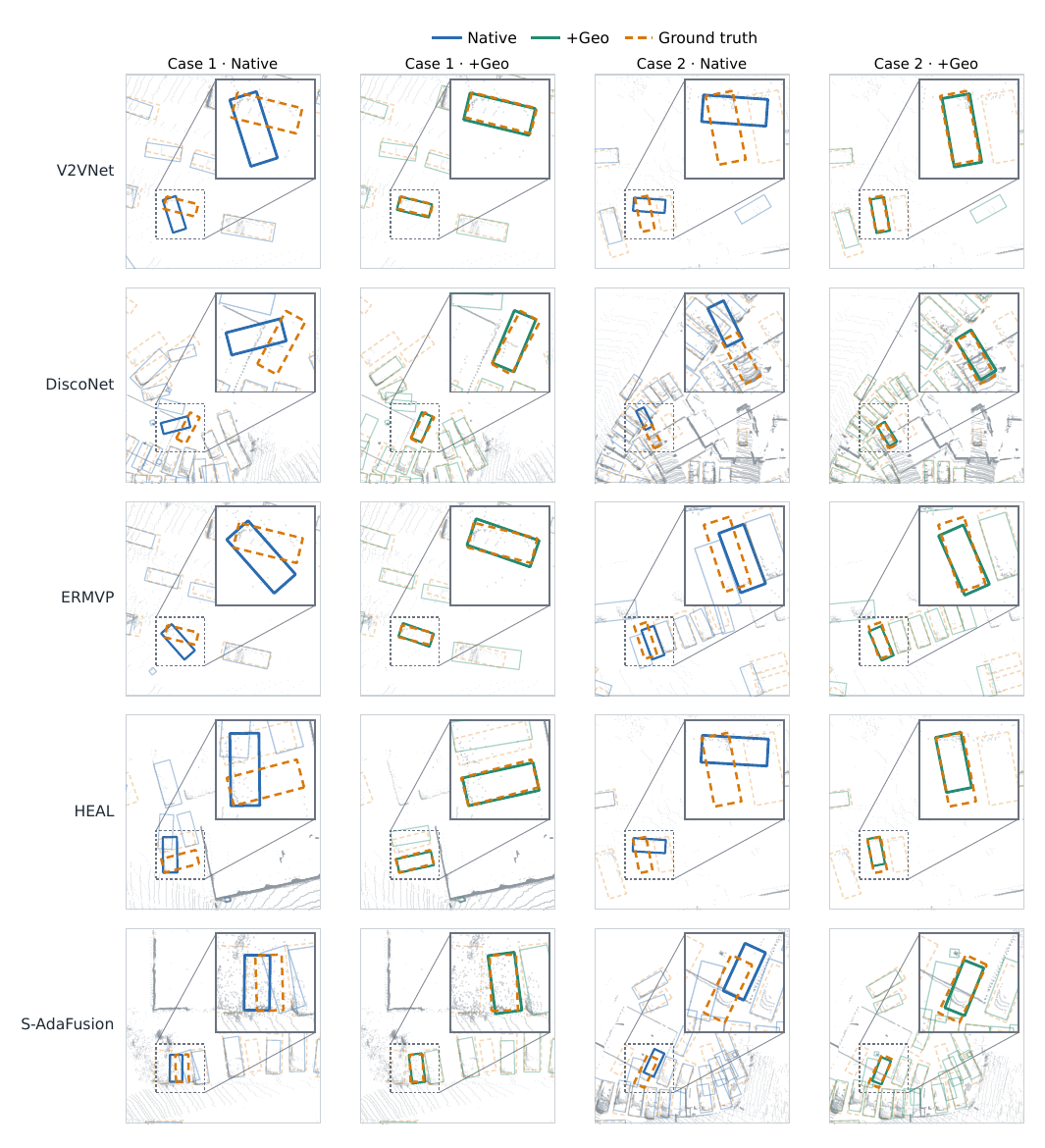}
    {Qualitative comparisons on crossing1/2 (continued)}
    {This complements Fig.~\ref{fig:qualitative-val-1}.}
    {fig:qualitative-val-2}
\qualitativepage{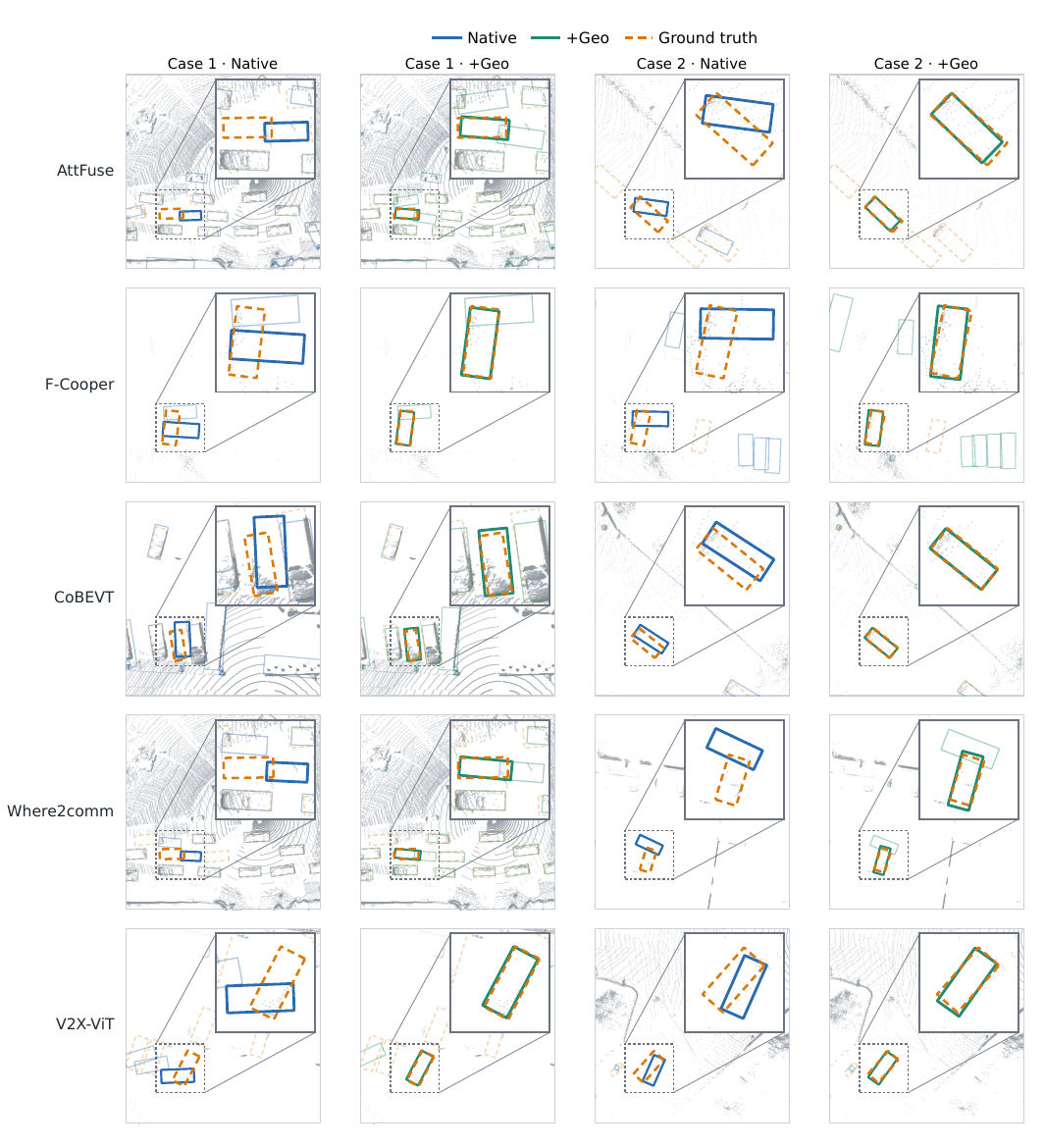}
    {Qualitative comparisons on crossing3}
    {The remaining methods are shown in Fig.~\ref{fig:qualitative-test-2}.}
    {fig:qualitative-test-1}
\qualitativepage{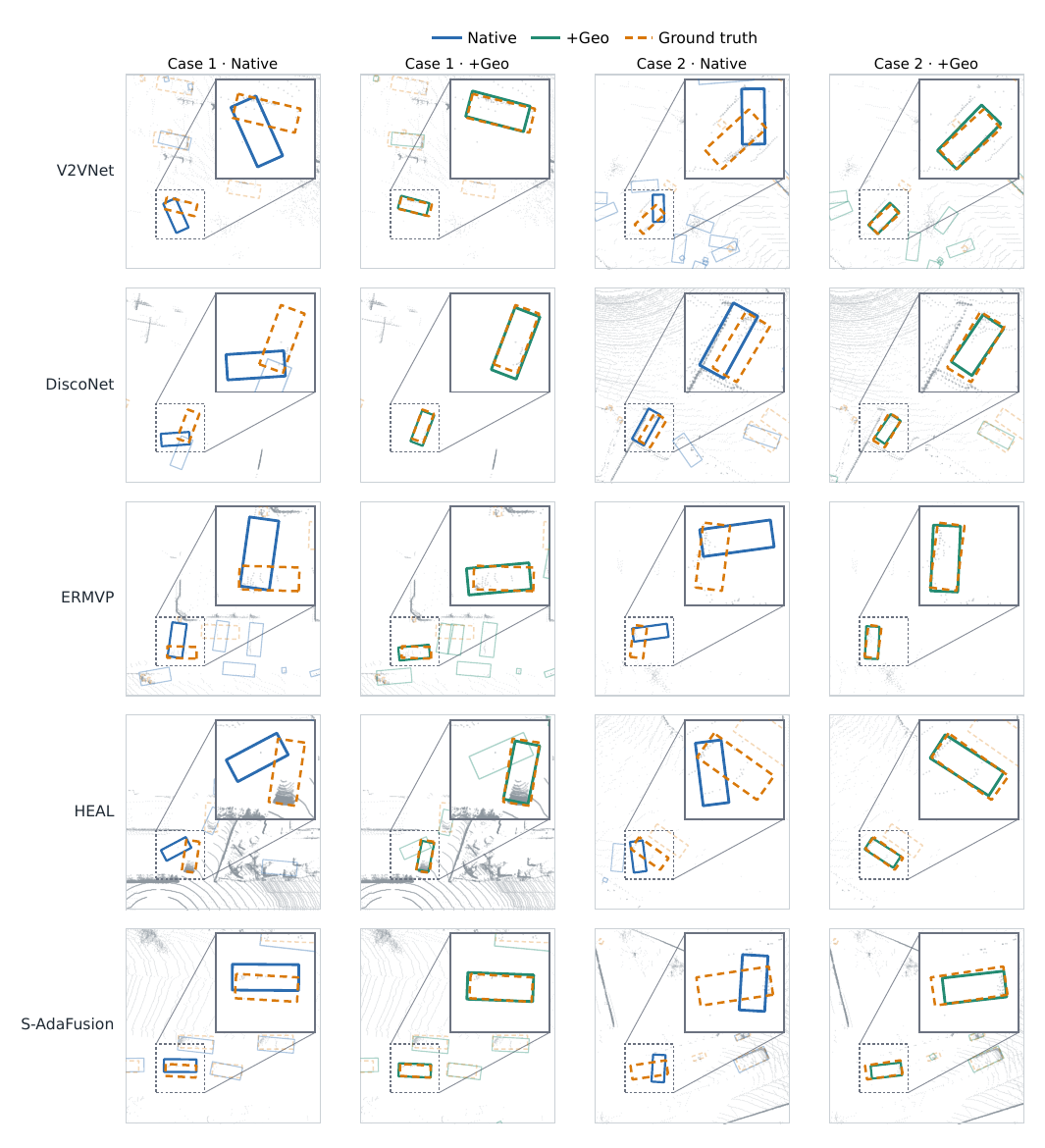}
    {Qualitative comparisons on crossing3 (continued)}
    {This complements Fig.~\ref{fig:qualitative-test-1}.}
    {fig:qualitative-test-2}
\end{document}